\pdfoutput=1
\documentclass[11pt]{article}

\usepackage[]{acl}

\usepackage{times}
\usepackage{latexsym}
\usepackage{multirow}
\usepackage{todonotes}

\usepackage[T1]{fontenc}
\usepackage[utf8]{inputenc}

\usepackage{microtype}
\usepackage{svg}
\usepackage{inconsolata}
\usepackage{multirow}
\usepackage{booktabs}
\usepackage{graphicx}
\usepackage{subfigure}
\usepackage{tcolorbox}
\usepackage{xcolor}
\usepackage{bbm}
\usepackage{amsmath}
\usepackage{comment}
\usepackage{caption}
\usepackage{xspace}

\newcommand{\dataset}{EXAMS-V\xspace}
\newcommand{\Ninstances}{20,932\xspace}

\newcommand{\Nlanguages}{11\xspace}
\newcommand{\Nlanguagefamilies}{7\xspace}
\newcommand{\Nsubjects}{20\xspace}
\newcommand{\NExam}{83\xspace}
\newcommand{\bul}{bg\xspace}
\newcommand{\chin}{zh\xspace}
\newcommand{\hrv}{hr\xspace}
\newcommand{\fra}{fr\xspace}
\newcommand{\deu}{de\xspace}
\newcommand{\hun}{hu\xspace}
\newcommand{\eng}{en\xspace}
\newcommand{\srb}{sr\xspace}
\newcommand{\itl}{it\xspace}
\newcommand{\ara}{ar\xspace}
\newcommand{\pol}{pl\xspace}
\title{\dataset: A Multi-Discipline Multilingual Multimodal Exam Benchmark for Evaluating Vision Language Models}

\author{
Rocktim Jyoti Das$^{1}$ \quad Simeon Emilov Hristov$^2$ \quad  Haonan Li$^{1}$ \\
\textbf{Dimitar Iliyanov Dimitrov}$^2$ \quad  \textbf{Ivan Koychev}$^2$ \quad  \textbf{Preslav Nakov}$^1$ \\
$^1$MBZUAI \quad $^2$Sofia University\\
\texttt{\{rocktimjyotidas, preslav.nakov\}@gmail.com}
}

\begin{document}
\maketitle
\begin{abstract}
We introduce \dataset, a new challenging multi-discipline multimodal multilingual exam benchmark for evaluating vision language models. It consists of  \Ninstances multiple-choice questions across \Nsubjects school disciplines covering natural science, social science, and other miscellaneous studies, e.g.,~religion, fine arts, business, etc. \dataset includes a variety of multimodal features such as text, images, tables, figures, diagrams, maps, scientific symbols, and equations. The questions come in \Nlanguages languages from \Nlanguagefamilies language families. Unlike existing benchmarks, \dataset is uniquely curated by gathering school exam questions from various countries, with a variety of education systems. This distinctive approach calls for intricate reasoning across diverse languages and relies on region-specific knowledge. Solving the problems in the dataset requires advanced perception and joint reasoning over the text and the visual content of the image. 
Our evaluation results demonstrate that this is a challenging dataset, which is difficult even for advanced vision--text models such as GPT-4V and Gemini; this underscores the inherent complexity of the dataset and its significance as a future benchmark.\footnote{Datasets are available at \url{https://github.com/RocktimJyotiDas/EXAMS-V}}
\end{abstract}

\section{Introduction}

\begin{figure}[t]
    \centering
    \hspace*{-1em}
    \includegraphics[width=1.1\columnwidth]{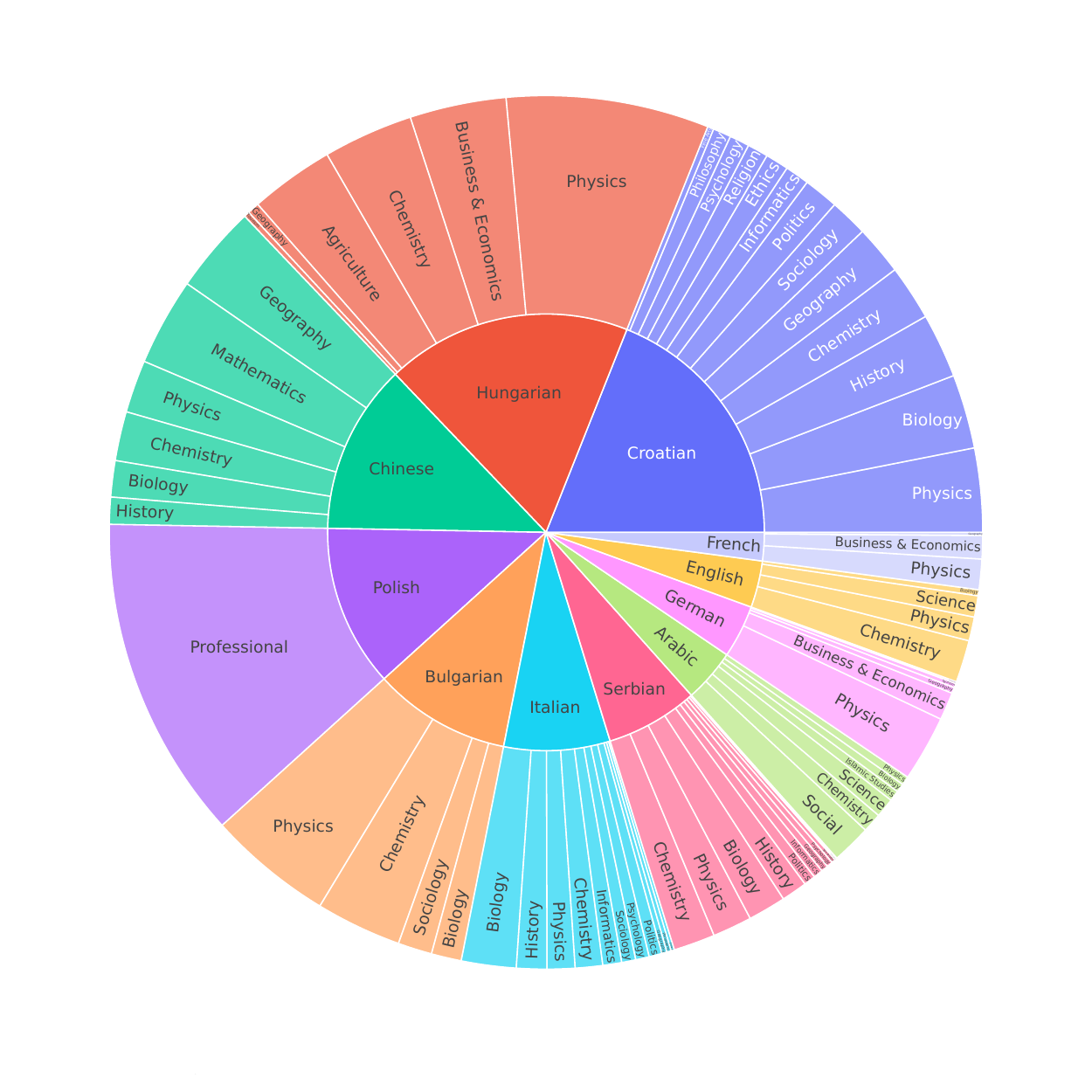}
    \vspace{-3em}
    \caption{Data distribution for our \dataset dataset: languages and subjects.}
    \label{fig:data-dist}
\end{figure}

Large Language Models (LLMs) have recently demonstrated impressive skills in understanding and generating natural languages \citep{Brown2020LanguageMA,opt,bloom,zeng2023glmb,Touvron2023Llama2O}. This progress has paved the way for significant advancements in LLM-based vision models \citep{zhu2023minigpt,liu2023llava}. Notable developments like GPT-4V \citep{openai2023gpt4} and Gemini \citep{Anil2023GeminiAF} represent a new era in image understanding, exhibiting remarkable proficiency in interpreting and analyzing visual data alongside textual information.
However, as Vision Language Models (VLMs) grow more sophisticated, existing benchmarks are becoming outdated, and unable to accurately assess these models' performance. 

For LLM evaluation, standardized testing akin to school examinations has proven to be an effective measure of a model's capabilities. 
A typical benchmark MMLU \citep{mmlu}, which contains 57 subjects across science, engineering, and humanities, has become a de facto benchmark for LLM evaluation. 
Several other school exam datasets have also set the standard in evaluating LLMs in different languages \citep{hardalov-etal-2020-exams,li2023cmmlu,koto2023large}.

In terms of VLM, a comparable benchmarking framework is conspicuously absent.
Existing benchmarks are (1) primarily monolingual, focused on English; (2) mostly not from school exams, leading to differences in methods of examining humans; (3) tend to keep images and text separate, which fails to challenge models with more complex tasks involving integrated visual elements like tables, symbols, and scientific notations. 
% English VLM benchmark \citep{VQA,hudson2019gqa,gurari2018vizwiz,lu2022learn,yue2023mmmu}

\begin{figure*}[t]
    \centering
    \includegraphics[width=1\textwidth]{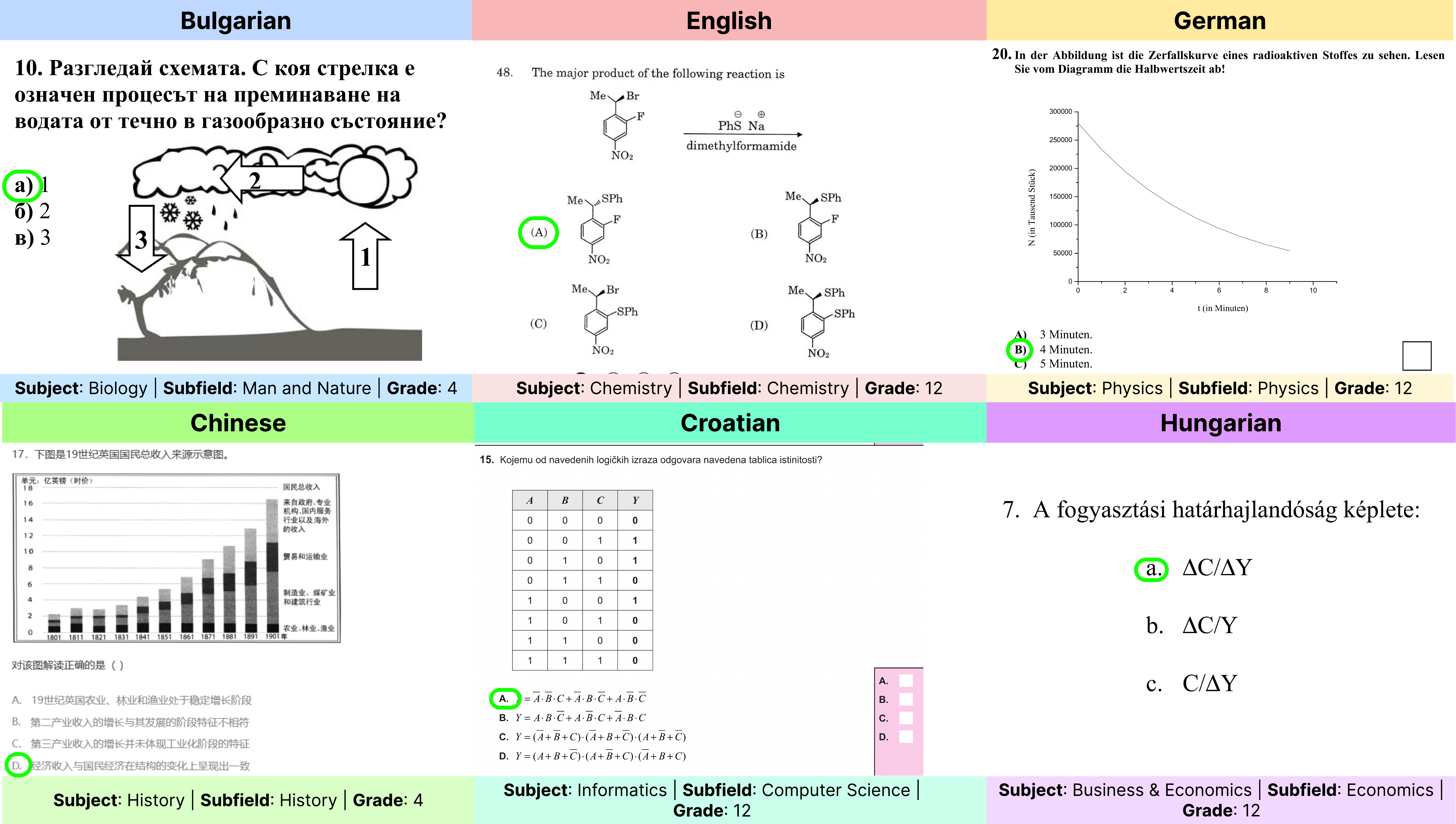}
    \caption{Sampled \dataset examples from different languages. The questions require the ability to understand multiple languages in addition to expert perception and reasoning capabilities.}
    \label{fig:low_data}
\end{figure*}

We introduce \dataset, which addresses all these issues. First, this dataset represents a significant leap forward, treating visual and text content as a cohesive unit. This forces models to engage in more sophisticated processing, including distinguishing, preprocessing, and logical reasoning over combined textual and visual information. 
Additionally, \dataset has a multilingual reach, covering \Nlanguagefamilies language families, further enhancing its complexity and applicability.

The key contributions of our paper include:
\begin{itemize}
\item We introduce a novel dimension to benchmarking vision language models, requiring them to reason over a unified snapshot that includes text, images, tables, graphs, and more. For this, we propose a new multimodal multilingual dataset, \dataset, comprising \Ninstances questions, spanning \Nlanguages languages and \Nsubjects subjects.
\item We evaluate the performance of state-of-the-art large language models and vision language models on our proposed dataset.
\end{itemize}

Through \dataset, we aim to set a new standard in evaluating VLMs, providing a more realistic and challenging benchmark that mirrors the complexity and diversity of real-world information processing.

\begin{table*}[t]
\centering
\begin{tabular}{lccc}
\toprule
\textbf{Dataset} & \textbf{Size} &  \textbf{Source}  & \textbf{Answer}  \\ 
\midrule
% VQA \citep{VQA} & \textgreater 1M & && No & Annotated & 1& Open\\ %\hline
% GQA \citep{hudson2019gqa} & \textgreater 1M & &&No & Synthesized & 1 & Open\\ %\hline
% VisWiz \citep{gurari2018vizwiz} & 32K &&& No & Annotated & 1 & Open\\ %\hline
% TextVQA \citep{singh2019vqa} & 45K & &&No & Annotated & 1 & MC\\ %\hline
% OKVQA \citep{marino2019okvqa} & 14K & &&No & Annotated & 1 & Open\\ %\hline
% SEED \citep{li2023seedbench} & 19K & &&No & Annotated & 1 & MC\\ %\hline
MMBench \citep{liu2023mmbench} &  2974 & Repurposed from 12 existing datasets &  MC\\ %\hline
MM-Vet \citep{yu2023mmvet} & 200  &  Internet images and annotated questions &  Open\\ %\hline
ScienceQA \citep{lu2022learn} & 21,198  & Textbooks& MC\\ %\hline
MMMU \citep{yue2023mmmu} & 11,550 & Textbooks, Internet, Annotated & Open/MC \\ %\hline
MathVista \citep{Lu2023MathVistaEM} & 6,141  & Repurposed from 28 existing dataset  & Open/MC\\ %\hline
M3Exam \citep{zhang2023m3exam} & 12,317 &  Exam Papers & MC\\ 
\midrule
\dataset & \Ninstances  & Exam Papers & MC\\ 
\bottomrule
\end{tabular} 
\caption{Comparison of \dataset with existing benchmarks. Here, "repurposed" means the benchmark is a compilation of prior datasets, MC refers to multi-choice type questions, and "open" refers to open-ended generation questions. }
\label{tab:data-comp}
\end{table*}

\begin{table}[]
\begin{tabular}{lcc}
\toprule
 & \multicolumn{1}{l}{\bf M3Exam} & \multicolumn{1}{l}{\bf \dataset} \\ %\hline
 \midrule
Interleaved & No & Yes \\ %\hline
Languages & 9 & 11 \\ %\hline
Min Sub. in a lang & 1 & 3 \\ %\hline
Max Sub. in a lang & 12 & 13 \\ %\hline
Avg. Sub. per lang  & 5 & 7.1 \\ %\hline
Samples & 12,317 & 20,946\\ 
Multimodal samples & 2,816 & 5,086 \\ 
\bottomrule
\end{tabular}
\caption{Comparison of M3Exams with EXAMS-V. Here, interleaved means that multimodal elements, like tables, figures, etc., are interleaved with the textual information in the image. The average subject per language for \dataset is reported by excluding Polish because Polish is a collection of 55 different professional exams that cannot be directly mapped to conventional subjects.}
\label{tab:m3exam_comp}
\end{table}

%\todo{revise the caption of figure 1}

\section{Related Work}
LLM witnessed remarkable advancements in recent years, enabling them to generate human-like text, answer complex questions, and perform a wide range of NLP tasks \citep{Brown2020LanguageMA,palm,touvron2023llama,vicuna2023,liu2023llm360}. 
Simultaneously to the rapid development of English-centroid LLMs, researchers have also focused on extending mono-lingual language models to multilingual \citep{bloom,zeng2023glmb,bactrian-x,jais} and multimodal \citep{alayrac2022flamingo,VisualGPT,liu2023llava,li2023blip2,Bai2023QwenVLAF}.
Models, such as GPT-4 \citep{openai2023gpt4}, Gemini \citep{Anil2023GeminiAF} have demonstrated exceptional performance on various benchmarks and have been widely adopted in academia and industry. 
However, the evaluation of these models is a critical aspect that requires careful consideration to ensure reliable and comprehensive assessments.

Several benchmarks have been proposed to assess the multimodal capabilities of LLMs \cite{VQA,hudson2019gqa,gurari2018vizwiz,singh2019vqa, lu2022learn, yue2023mmmu, Lu2023MathVistaEM}. Most early-stage benchmarks consist of photos as images, and the questions ask about the objects, attributes, or relationships between objects in the image. 

Recently, inspired by the use of school exams as benchmarks for LLMs, researchers have begun to collect curriculum-based questions with images for VLM benchmarking. ScienceQA \citep{lu2022learn} is one of the most popular datasets in this area. 
It contains 21,208 multimodal multiple-choice questions with rich domain diversity across 26 topics, collected from elementary and high school science curricula. To answer these science questions, a model needs to understand multimodal content and extract external knowledge to arrive at the correct answer.
MMMU \citep{yue2023mmmu} is another benchmark designed to evaluate multimodal models on massive multi-discipline tasks demanding college-level subject knowledge and deliberate reasoning. It includes 11,550 questions from college exams, quizzes, and textbooks, covering six core disciplines: art, business, science, health, humanities, and technology.  Similarly, MathVista \cite{Lu2023MathVistaEM} is a benchmark with 6,141 samples for evaluating the mathematical reasoning capabilities in a visual context.
However, like all previous benchmarks, these two exam benchmarks are in English.

M3Exam \citep{zhang2023m3exam} is the first multilingual multimodal exam benchmark that covers 9 languages. It includes 12,317 questions, with 2,816 questions requiring information from an image to arrive at the answer. 
One main difference between M3Exam and our dataset is that, like all other VLM benchmarks, M3Exam separates text and images for a single question, while we embed the question in the images.
% In a realistic scenario, the model is expected to possess advanced perception capabilities to extract the relevant text information from the image, including parsing tabular data, graphs, scientific formulas and equations, along with the ability to do multimodal reasoning. Additionally, 
% Furthermore, among the nine languages under consideration, three of them have only two subjects, and among these, two languages exclusively consist of language comprehension questions. This compromises the comprehensive evaluation of these languages.
% 23.5 percent of the questions in the dataset are language comprehension or language understanding questions without any image reference. These questions do not require explicit knowledge of a particular discipline or complex reasoning. 

Unlike the above benchmarks, our dataset boasts a broader linguistic scope, placing a particular emphasis on low-resource languages like Croatian, Hungarian, Spanish, and French. Notably, our examination benchmark surpasses others by featuring a greater number of questions, encompassing a diverse range of types and topics. This variety includes questions with accompanying images, tables, and graphs, as well as mathematical and chemistry equations. For a detailed quantitative analysis, please refer to Table~\ref{tab:data-comp}.

% \noindent{\bf Textbook QA} The Textbook QA dataset raises a new question in the context of visual question answering. It proposes the task of Multi-Modal Machine Comprehension (M3C), an extension of the traditional textual machine comprehension to multi-modal data. In this paradigm, the task is to read a multi-modal context along with a multi-modal question and provide an answer, which may also be multimodal in nature. This is in contrast with the conventional question-answering task,in which the context is usually about a single modality. The TQA dataset consists of 1,076 lessons with 26,260 multi-modal questions.\\

\section{\dataset Dataset}

\dataset is a multimodal extension of the EXAMS dataset \citep{hardalov-etal-2020-exams}, which is collected from official state examinations crafted by the ministries of education across different countries. These assessments, taken by high school graduates, cover diverse subjects, including core disciplines like Biology, Chemistry, Geography, History, and Physics, as well as specialized areas such as Economics and Informatics. The original EXAMS dataset was intended for multilingual question answering and thereby ignored the questions requiring visual information. We included additional data for English and Chinese in our \dataset dataset. 
The subject coverage and statistics are detailed in Table \ref{tab:stat_data2}.

\subsection{Data Collection and Analysis}

\paragraph{Collection and Preparation of Dataset.}
% The dataset collection process consisted of three steps. First, we identified potential online sources of publicly available school and entrance exams.\todo{Rocktim give examples of the data sources. If only a few sources - list them. If not put them in a table or graph.} These exams are taken by students graduating from high school. We downloaded the PDF files per year, per subject, from the identified sources. We adhere to copyright and license regulations, avoiding data from sites prohibiting copy and redistribution. 
% Given the arising data contamination concerns of foundational models, we only collected exams which are available in PDF formats.
The dataset collection process consisted of three main steps. Initially, we retraced the original PDFs used for creating the original EXAMS dataset. For English and Chinese, we gathered high school and entrance exam questions (specifically, Gaokao and JEE Advanced Questions) from China and India, respectively.

Then, we converted each PDF document to a series of cropped images, with each image having a single question and possible answers with accompanying tables, images, graphs, etc. This required the conversion of each page in the PDF document to an image and then the use of an open-sourced labeling pipeline to place bounding boxes around each question and its answers for each page.\footnote{\url{https://github.com/Cartucho/OpenLabeling}}

The third step involved the creation of metadata for each cropped question. This metadata includes a unique ID, file path to the question snapshot, subject, grade, language, and the correct answer for the question. Each set of metadata is stored as a JSON file corresponding to a specific subject in a particular language.

\paragraph{Annotation Guidelines.} All the bounding box annotations are done manually by the authors with the following agreed-up guidelines: Only multiple-choice questions with 3 to 5 options and exactly \textbf{one} correct answer are considered, as they allow for a standard automatic evaluation of the correctness of model outputs;
%\begin{itemize}
    % \item We exclude subjective questions with free-response answers because of the difficulty of automatic evaluation;
    % \item We also remove any true-false statements questions to maintain the difficulty of the dataset.
%\end{itemize} 

Along with placing the bounding boxes, we marked whether the context within the bounding box is pure text or has visual context like table, graph, figure, or symbols. 
As the result of annotation, each question sample is an image that contains the question text and candidate options, along with other vision information such as figures, tables, graphs, etc. 
It also includes meta-information, as mentioned beforehand. This rigorous process allowed us to maintain the high quality of the dataset.
% \begin{table}[]
% \centering
% \begin{tabular}{lcccc}
% \toprule
%  & IC & QC & SC & Others \\ \midrule
% \bul & 100.00 & 100.00 & 100.00 & 100.00 \\ %\hline
% \srb & 100.00 & 100.00 & 100.00 & 100.00 \\ %\hline
% \hrv & 100.00 & 100.00 & 100.00 & 100.00 \\ %\hline
% \itl & 100.00 & 100.00 & 100.00 & 100.00 \\ %\hline
% \chin & 98.00 & 98.00 & 100.00 & 100.00 \\ %\hline
% \eng & 100.00 & 100.00 & 100.00 & 98.00 \\ %\hline
% \ara & 100.00 &100.00&100.00& 100.00 \\ \bottomrule
% \end{tabular}
% \caption{Data Quality Analysis Results}
% \label{tab:data_quality}
% \end{table}
\paragraph{Data Quality Assessment.}After the completion of our annotation process, we conducted a data quality assessment on seven languages based on the availability of an annotator with language expertise. In this evaluation, we randomly selected 50 questions from each language and requested annotators to assess each image sample based on four binary criteria:
\begin{itemize}
    \item Image Clarity: Clarity of visual elements such as images, diagrams, or tables.
    \item Question Clarity: Clarity of textual information in the question.
    \item Single Correct: The image contains a single Multiple Choice Question (MCQ) with precisely one correct option.
    \item Others: Identification of other issues. The other issues encompass factors that render a question invalid, such as the presence of the answer within the question snapshot.
\end{itemize}

A question is deemed completely valid only if it meets all four criteria.

Upon thorough review, all annotators unanimously deemed the samples to exhibit exceptionally high quality across all annotated languages. Specifically, all the samples in Bulgarian, Croatian, Serbian, Italian, and Arabic met the four quality assessment criteria. However, in the case of Chinese, one sample exhibited unclear image information, and another displayed a question that lacked clarity. Similarly, an English sample was deemed invalid due to the presence of the answer within the image sample. This proves the high quality of our dataset. The annotation guideline used by the annotators is provided as Figure \ref{fig:guideline} in the Appendix section \ref{app:guid}.

% \todo[inline]{Table 3: Put number of languages (instead of multilingual), average subjects/language, truly multimodal examples, number of subjects, language families, has parallel questionsq.}
\begin{table*}[t]
\begin{tabular}{lccccccc}
\toprule
\textbf{Language} & \textbf{ISO} & \textbf{Family} & \multicolumn{1}{c}{\textbf{Grade}} & \multicolumn{1}{c}{\textbf{\# Subjects}} & \multicolumn{1}{c}{\textbf{\# Questions}} & \multicolumn{1}{c}{\textbf{\# visual Q.}} & \multicolumn{1}{c}{\textbf{\# text Q.}} \\ 
\midrule
English 	&  \eng& Germanic & 11, 12 	& 4 	& 724 	& 181 	& 543  \\
Chinese 	& \chin  & Sino-Tibetan 	& 8-12 	& 6 	& 2,635 	& 1,991 	& 644 \\ 
French 	    & \fra & Romance 	    & 12 	& 3 	& 439 	& 50 	& 389 \\ 
German 	    & \deu & Germanic 	    & 12 	& 5 	& 819 	& 144 	& 675 \\ 
Italian 	& \itl & Romance  	    & 12 	& 11  	& 1,645 	& 292 	& 1,353 \\
Arabic 	    & \ara & Semitic 	    & 4-12 	& 6 	& 823 	& 117 	& 706 \\  
Polish 	    & \pol & Slavic	& 12 	& 1 	& 2,511  	& 422 	& 2,089 \\
Hungarian 	& \hun & Finno-Ugric 	& 12 	& 6 	& 3,801 	& 495 	& 3,306 \\ 
Bulgarian 	& \bul & Slavic 	& 4, 12 	& 4 	& 2,132 	& 435 	& 1,697 \\ 
Croatian 	& \hrv & Slavic	& 12 	& 13 	& 3,969 	& 700 	& 3,269 \\
Serbian 	& \srb & Slavic	& 12 	& 11 	& 1,434 	& 259 	& 1,175 \\

\bottomrule
% Romanian & Romance & 12 & 1 & 5 & 0 & 5 \\ %\hline
% Russian & East-Slavic & 12 & 1 & 9 & 0 & 9 \\ %\hline
% Slovakian & West-Slavic & 12 & 1 & 46 & 4 & 42 \\ %\hline
% Spanish & Romance & 12 & 2 & 299 & 89 & 210 \\ \hline
\end{tabular}
\caption{Statistics of \dataset dataset. The languages are ordered from high-resource to low-resource languages. Here, $\#$ visual Q. refers to questions with multimodal context and $\#$ text Q. refers to text only questions.}
\label{tab:stat_data}
\end{table*}

\subsection{Data Statistics}
The \dataset dataset contains \Ninstances samples in total, spanning \Nsubjects subjects from grade 4-12. It encompasses a total of \Nlanguages languages from \Nlanguagefamilies language families, while it contains parallel data in more than three languages.\footnote{Parallel data means that questions are semantically the same but in different languages.}
The statistics are presented in Table~\ref{tab:stat_data}, and Table~\ref{tab:stat_data2} per language per subject details.

\paragraph{Language Diversity.} 
Table \ref{tab:stat_data} provides an overview of the various languages featured in the dataset, along with the number of questions and subjects available for each language. 
The dataset includes high-resource languages like English and Chinese and low-resource languages such as Bulgarian, Croatian, and Serbian. 
It offers a diverse linguistic landscape, spanning Germanic, Slavic, and Sino-Tibetan language families. We also include Arabic, which has a script directionality from right to left. 
Additionally, Slavic and Romance language families exhibit multiple language representations, enabling the evaluation and understanding of closely related languages. These characteristics make \dataset a great fit for multimodal multilingual assessment of any LLMs and VLMs.

\paragraph{Parallel Questions.} Examinations in Croatia and the United Arab Emirates are administered in multiple languages, facilitating the development of parallel question sets for two language groups. 
Specifically, for Croatian examinations, we have parallel questions available in both Serbian and Italian. Additionally, Arabic questions are paired with English counterparts for four subjects: Science, Physics, Chemistry, and Biology. 
This process resulted in the creation of 1,207 Serbian questions and 1,147 Italian questions in parallel with Croatian. Furthermore, for Arabic, we have developed 262 parallel questions in English.

\paragraph{Subject Diversity.}
Each education system has its own specifics, leading to some differences in curricula, topics, and even the naming of the subjects. 
As a result, we initially collected \NExam different subjects from different countries. Since different naming conventions for subjects in different countries, the values of the subjects were very sparse and non-uniformly populated. We performed subject aggregation to club similar subjects into one single subject and finally got \Nsubjects aggregated subjects. 
They were further grouped into three major categories, based on the main branches of science: Natural Sciences -- the study of natural phenomena; Social Sciences -- the study of human behaviour and societies; others -- Applied Studies, Arts, Religion, etc. 
The distribution of the major categories is Natural Sciences (53.02\%), Social Sciences (27.15\%), and Others (19.82\%).

\paragraph{Question Complexity.}
The dataset is compiled from high school examinations administered in various countries, primarily featuring questions from grades 4 to 12. 
Questions in natural sciences such as Physics, Chemistry, Biology, and Mathematics, demand foundational knowledge of these subjects and intricate reasoning skills. 
Questions related to Geography and History necessitate specific knowledge about particular regions or countries. 
Additionally, the Polish section comprises a compilation of 55 diverse professional exam questions across various fields, spanning from accounting to the motor vehicle service process. Answering these questions requires precise understanding of these professions. 

\subsection{Comparison with Existing Datasets}

\dataset differs from other datasets by mainly introducing a new way of benchmarking VLMs -- passing an entire question snapshot that contains both the visual and the text components instead of passing the parsed and processed text with the image. This leaves the model to the work of text extraction and representation. 
Moreover, the dataset has questions of varying complexity and diversity with most of the questions coming from high school matriculation exams. Most previous benchmarks normally require commonsense knowledge or simple physical or temporal reasoning. In contrast, the \dataset benchmark requires deliberate reasoning with high school-level subject and region-specific knowledge. 
Lastly, \dataset aims to cover high school-level knowledge with different forms of visual features, including diagrams, tables, charts, chemical structures, paintings, geometric shapes, etc. This means that a well-performing model on \dataset could be considered to surpass a human adult on general-purpose tasks. 
We have included a detailed comparison of \dataset dataset with other benchmarks in Table \ref{tab:data-comp}.

\begin{table*}[t]
\centering
\scalebox{0.85}{%
    \begin{tabular}{lcccccccccccc}
    \toprule
    \textbf{Model} &\textbf{ \bul }& \textbf{\chin} & \textbf{\hrv} & \textbf{\fra} & \textbf{\deu} & \textbf{\hun} & \textbf{\eng}& \textbf{\srb} & \textbf{\itl} & \textbf{\ara} & \textbf{\pol} & \textbf{Avg}  \\ \midrule 
    Random &25.23& 24.13 & 25.58&24.14&22.56&19.55&24.40&24.50&24.76& 19.83&  23.00& 23.62 \\ 
    % Frequent Choice &&  &  &&&&&&&& &   \\
    \midrule
    \multicolumn{12}{c}{\textbf{Vision Language Models (VLMs)}} \\ 
    \midrule
    LLaVA-1.5-13B &--&--   &-- &--&--&--& 26.00&--&--&--& --&--\\ 
    Qwen-VL-7B &--&   15.72&-- &--&--&--&23.60&--&--&--& --&-- \\ 
    GPT-4V& \textbf{36.00}& 22.20  & 55.47 &\textbf{60.34}&\textbf{51.24}&\textbf{44.77}&29.27&39.84&62.07& 24.29& 30.00&42.78\\
    % Bard &52.92& 37.33  & 50.48&48.45&59.49&53.33&31.20\\ %\hline
    Gemini-V &30.46& \textbf{24.56}&29.39&\underline{47.70}&\underline{47.80}&27.05&29.20&28.29&43.03& 19.38 & 28.00&31.13 \\ 
    \midrule
    \multicolumn{12}{c}{\textbf{Augmented Large Language Models (LLMs): OCR + Captioning} }\\ 
    \midrule
    GPT-3.5 Turbo &27.08 & 22.20 & 52.08 &39.08&34.81&37.73&30.00&48.61&55.48& \underline{26.36} & \underline{33.00}& 39.47\\
    GPT-4 &30.46 & 23.57 &\textbf{66.58}&36.71&23.76&34.09&\textbf{32.40}&\textbf{73.51}&\textbf{75.95}& \textbf{26.47}& 30.00& \textbf{47.11} \\
    Gemini Pro & \underline{32.00}& \underline{23.97}& \underline{58.90} &38.51&28.09&\underline{43.41}& \underline{31.20} &\underline{59.96}&\underline{64.38}& 23.25& \textbf{42.00}&\underline{43.99}\\ 
    \bottomrule
    \end{tabular}
}
\caption{Overall results for different models on \dataset test set. Besides reporting performance for VLMs, we additionally add text-only LLM baselines. The best-performing model in each category is in bold, and the second-best is underlined.}
\label{tab:model-comp}
\end{table*}

\section{Experimental Setup}

As we see in Table~\ref{tab:stat_data2}, the original data appears sparse and imbalanced. To ensure a more balanced benchmark, we split \dataset into training and test sets, with careful consideration for language and subject representation in the test set.
We sampled 20 to 100 questions for each subject-language pair based on availability. For languages with parallel data like Croatian, Serbian, and Italian, we performed parallel splits to maintain question consistency across training and test sets.
Finally, we got 16,724 training and 4,208 test instances.

We evaluate state-of-the-art LLMs and VLMs on \dataset benchmark.
Our evaluation is conducted under a zero-shot setting without model finetuning or in-context learning, using either APIs or NVIDIA A100 GPUs.\footnote{Our experiments were conducted in Dec-2023 and attached to the latest version of the commercial models.}

\subsection{Models}
\paragraph{VLMs.} We consider various large vision language models. We evaluated two open-source models, which have shown remarkable performance on multiple multimodal tasks: (i) LLaVA-1.5 \cite{liu2023llava} which integrates visual embeddings with Vicuna's linguistic space. (ii) Qwen-VL-Chat \cite{Bai2023QwenVLAF}, a multilingual multimodal chat model trained on Chinese and English data, which possesses excellent grounding, text-reading, and text-oriented question-answering performance.  
We also evaluated two proprietary multimodal models: GPT-4V and Gemini-Pro-Vision (denoted as Gemini-V) \cite{Anil2023GeminiAF}. 
GPT-4V is the best-performing multimodal model by OpenAI, and Gemini-V is the mid-range model among the Gemini family of multimodal models.

\paragraph{Augmented LLMs.} To evaluate text-only LLMs, we augment language models with two image-to-text tools, namely Optical Character Recognition (OCR) and Image Captioning (IC). 
We employ Google Tesseract for OCR and GPT-4V for image captioning. We treat LLM augmented with OCR and IC as a vision language system. This setup was applied to GPT-3.5-Turbo, GPT-4, and Gemini Pro.

\subsection{Evaluation Setup}  Given the multiple-choice nature of the questions, accuracy served as our primary metric. Models were instructed to format answers as JSON objects \{"answer": "choice"\}, allowing for straightforward prediction extraction from the outputs. Based on our observation, all models under consideration can adhere to the instructions and produce the answer in a JSON format.

% \todo[inline]{Add results with VLM+OCR; and maybe VLM+OCR+Captioning.}

% \todo[inline]{Put a detailed table of results for language--subject combination.}

% \todo[inline]{Try new LLaVA-1.6.}

% \todo[inline]{About Table 3: you need in the appendix, examples of what goes as input to the LLM/VLM. So that people 
%  can see the OCR and the captioning for a specific (English) example.}

%  \todo[inline]{Try also VLM that has as input not just the image, but also the OCR + Caption.
%  Also try the OCR + Captioning translated to English (as a separate experiment).}

%  \todo[inline]{Try also GPT4 that has as input not just the OCR + Caption, but also the textual analysis from the VLM. Also try the OCR + Captioning translated to English (as a separate experiment).}

% \todo[inline]{You also need to say the exact number of test examples for each table: Table 3 and 4 -- give the number of examples for each language, and also overall.}

\begin{table*}[h]
\centering
\begin{tabular}{lccc|ccc|ccc}
\toprule
\multirow{2}{*}{\textbf{Subject}} &  \multicolumn{3}{c|}{\textbf{GPT-4V}} & \multicolumn{3}{c|}{\textbf{Gemini-V}}&  \multicolumn{3}{c}{\textbf{GPT-4 (w/ OCR, captions)}} \\
\cmidrule{2-10} 
  & \textbf{\hrv} & \textbf{\srb} & \textbf{\itl} & \textbf{\hrv} & \textbf{\srb} & \textbf{\itl} & \textbf{\hrv} & \textbf{\srb} & \textbf{\itl} \\
\midrule
Biology	& \textbf{72.04}	& 37.64	& 66.90	& 32.26	& 31.18	& \textbf{43.41}	& 75.27	& 73.11	& \textbf{77.55}\\
Chemistry	& 48.00	& 28.00	& \textbf{53.33}	& 25.33	& 26.67	& \textbf{34.67}	&\textbf{ 72.00}	& 68.00	& 72.00\\
History	& 59.26	& 45.68	& \textbf{61.73}	& 29.63	& 23.46	& \textbf{37.03}	& \textbf{85.19}	& 77.78	& 76.54\\
Informatics	& 38.39	& 33.33	& \textbf{40.74}	& 42.59	& 33.34	& \textbf{46.29}	& 34.00	& 57.41	& \textbf{66.67}\\
Politics	& \textbf{82.22}	& 46.67	& 73.33	& 46.67	& 31.11	& \textbf{64.44}	& \textbf{97.78}	& 91.11	& 86.67\\
Psychology	& 85.19	& 55.56	& \textbf{88.89}	& 33.33	& 29.63	& \textbf{59.26}	& 92.59	& \textbf{100.00}	& 92.59\\
Sociology	& \textbf{63.33}	& 53.33	& 60.00	& 33.33	& 30.00	& \textbf{56.67}	& \textbf{80.00}	& 73.33	& 70.00\\
\midrule
Average	& \textbf{62.56}	& 40.44	& 62.11	& 33.33	& 28.76	& \textbf{45.25}	& \textbf{80.53}	& 75.29	& 76.60\\
\bottomrule
\end{tabular}
\caption{Fine-grained subject-wise comparison on the parallel Croatian--Serbian--Italian examples. For a particular VLM or augmented LLM, the best-performing language for each subject among the three languages is in bold. }
\label{tab:subject-comparison}
\end{table*}

\section{Main Results}

We present the results across languages in Table \ref{tab:model-comp}. To gain a clearer understanding of model performance, we establish a random baseline by assigning an option randomly from the available choices for each question. The random baseline for all languages ranges between 19-26\%.

\paragraph{VLM Results.} Among the various VLMs, GPT-4V stands out with the highest performance, achieving an overall average score of 42.78\%. This score, being only 20 percentage points above the random baseline, indicates significant potential for improvement in VLM capabilities. Gemini-V, following GPT-4V in our evaluation, achieves an overall average of 31.13\%.

In comparison to commercial VLMs, open-source VLMs such as LLaVA-1.5-13B and Qwen-VL-7B fall short in terms of language support and model performance. According to our findings, open-source VLMs are limited in language support (2 for Qwen and 1 for LLaVA) and their performance in these languages is close to the random baseline. On the other hand, commercial models exhibit broader language support, as evidenced by their performance surpassing random outcomes in almost all languages.

\paragraph{LLMs Augmented with OCR and Captioning.} Large language models enhanced with OCR and image captioning show superior average performance compared to standalone vision-language models. 
GPT-4, when augmented with OCR and captioning, demonstrates the highest overall performance among both VLMs and LLMs. This can be attributed to the precise OCR capabilities of Google Tesseract, the detailed captions produced by GPT-4V, and GPT-4's robust textual reasoning abilities. Furthermore, unlike prompt GPT-4V to directly generate the answer, augmented GPT-4 decouples the difficulties in visual information extraction and text reasoning.

\subsection{Analysis from a Language Perspective}
Comparing model performance in different languages, we find that all models show random-level results for Chinese (zh), which might be due to the inherent challenges associated with the Chinese subset. The Chinese subset, derived from the Gaokao exam, contains the highest proportion of vision features such as figures, tables, or graphs. This makes it difficult not only for single VLM but also for OCR and image captioning techniques to capture the visual information in text form fully. 

Following Chinese, Arabic (ar) and English (en) emerge as the next most challenging languages. For Arabic, the low performance is associated with the image sample itself. Figure \ref{fig:ar} shows an image sample of Arabic where we can see that, unlike other subjects, the answer choices do not have any letter associated with it. Thus, when the evaluated VLMs and LLMs are instructed to return the answer in the form of A, B, C or D, they find it very difficult to pinpoint the correct option.

For English, the top-performing models only scored about 8 $\%$ above the random baseline. The difficulty in English might be attributed to its sourcing from the Joint Entrance Exam (JEE) conducted every year for admission to Engineering Institute in India. To solve these questions, the model needs to be able to demonstrate complex multi-step reasoning along with a very good understanding of fundamental science- Physics, Chemistry, and mathematics.  

Overall, the best performing VLM, i.e. GPT-4V, outperforms in four languages, whereas LLMs with OCR and captioning capabilities excel in several languages. These four languages include Bulgarian (\bul), French (\fra), German (\deu) and Hungarian (\hun). If we refer to the test set distribution reported in Table \ref{tab:test_dist} of the Appendix, we observe that most of the samples in these languages have very few multimodal questions. Additionally, they have very few graphical and tabular questions on which GPT-4V tends to show poor performance according to Table \ref{tab:mul-eval}.   Other languages like Croatian (\hrv), Serbian (\srb), Italian (\itl), and Polish (\pol) have a fair distribution of multimodal and textual questions.

% \todo[inline]{Further analysis is needed to understand the reasons behind these patterns, which may relate to the varying levels of multimodal information across languages.}

\subsection{Parallel Data Evaluation} 
Since Croatian, Serbian, and Italian data come from the same examination, we conducted a parallel sample experiment for these languages. The GPT-4V, Gemini-V, and augmented GPT4 results are reported in Table \ref{tab:subject-comparison}. The results show dependence on the language for all the models under consideration. Augmented GPT4 has the least performance variance. This can be attributed to the accurate OCR capabilities of Google-Tesseract. 

For GPT-4V, there is a significant performance gap between Croatian and Serbian with Croatian outperforming Serbian by 20.12 $\%$. Although both languages are very similar and often mutually intelligible, their scripts differ significantly. Serbian is Cyrillic, whereas Croatian is Latin. Latin script is more widely used, and the majority of the most spoken languages in the world have Latin script. This can be attributed to the strong performance of GPT-4V in languages with Latin script, like Croatian and Italian. 

Even for Gemini-Vision-Pro
, there is a gap in performance between Croatian and Serbian.  The accuracy for Croatian is better than Serbian by 4.57$\%$. Gemini-V exhibits a notable performance disparity between Italian and Croatian, as well as Serbian. This discrepancy is likely because Italian, as a high-resource language, enjoys greater representation within the Gemini family of models.

% One advantage of \dataset dataset is that it encompasses questions from different subjects. Here, we examine the performance of VLMs and LLMs on questions from different subjects. The results are summarised in Table \ref{tab:subject-comparison}. We considered two VLMs, namely GPT-4V and Gemini and LLMs augmented with OCR and Captioning, namely GPT-3.5 and GPT-4V. We have considered two languages for our evaluation: Chinese and Croatian, which jointly cover all the subjects on which we are evaluating in our dataset. Additionally, one is a high-resource language, and the other is a low-resource language. Extending our previous observation, for Croatian, GPT-4V shows the best performance for all subjects. Meanwhile, for the Chinese, there is no consistent trend. However, for most of the subjects, Gemini-Vision-Pro shows the best performance. Additionally, in terms of subjects, the models tend to have strong performance in subjects such as Chemistry and physics that require complex reasoning. Meanwhile, for knowledge-based subjects such as biology and geography, the model's performance is comparatively worse. 

\subsection{Vision Feature Evaluation}
We compared the performance of GPT-4V and Gemini-V for four different vision features: scientific symbols, figures, graphs, and tabular data. We compared it against image samples with only textual information. For the evaluation, we curated samples of different vision features from the Croatian subset. The results are reported in Table \ref{tab:mul-eval}. 

GPT-4V shows fairly good performance for questions that involve scientific symbols and figures. However, it demonstrates poor performance for questions with graphs and tabular.  Surprisingly, Gemini-V can show better performance for tabular data when compared to GPT-4V. Nevertheless, its performance across the remaining three vision features --- scientific symbols, figures, and graphs, was subpar. 

\begin{table}[]
 \centering
\begin{tabular}{lccc}
\toprule 
{\bf Feature} & \multicolumn{1}{l}{\bf Samples} 
 & \multicolumn{1}{l}{\bf GPT-4V} & \multicolumn{1}{l}{\bf Gemini-V} \\ 
 \midrule
 Symbol &36& 52.78 & 25.00 \\ 
Figure &50& 60.00 & 22.00 \\ 
Graph &50& 42.00 & 26.00 \\ 
Table &40& 27.50 & 37.50 \\ 
Text &50& 62.00 & 48.00 \\
\bottomrule
\end{tabular}
\caption{Model performance on different vision features.}
\label{tab:mul-eval}
\end{table}

% \section{Further Analysis}
% \paragraph{Disparity between Open-source and Propriety VLMs.} Leading open-source models such as LLaVA-1.5-13B and Qwen-VL-7B find it very difficult to perform on our benchmark. Whereas proprietary models such as GPT-4V and Gemini outperform them by a significant margin. Additionally, LLaVA-1.5 is trained only on English data, whereas Qwen-VL-7B is trained only in English and Chinese. We have reported results on these models only in their respective trained languages. This significant difference in multilingual capabilities and benchmark performance indicates a gap in the capabilities of current open-source models compared to proprietary ones like GPT-4V.

% \paragraph{Comparison of OCR and caption-augmented LLM with VLM.} Table \ref{tab:model-comp} shows our the performance of VLMs and LLMs augmented with OCR and caption. Our results show that OCR and captioning do not help language models achieve the performance of multimodal models. This finding suggests that the \dataset benchmark requires models to contextual interpretation of visual information in terms of text information. This underscores the complexity of the multimodal tasks the benchmark presents.
% % \noindent{\bf Comparision across different types of multimodality.}

\section{Conclusion and Future Work} 
% \section{Conclusion}
The development of \dataset as a benchmark for assessing the multilingual and multimodal capabilities of VLMs marks a significant milestone in the journey towards multilingual models. Furthermore, the \dataset introduces a new dimension to visual question answering where the textual information is a part of the image. Thus, \dataset not only tests the multimodal reasoning capability of current VLMs but also their ability to do OCR in a multilingual context. This requires a strong perception capability to draw boundaries between textual questions and multimodal contexts like tables, figures, graphs, etc. Furthermore, the questions with in-depth knowledge of multiple disciplines or subjects and questions from physics, chemistry, and mathematics require intricate reasoning. These features collectively contribute to the considerable complexity of the \dataset. We believe the evaluation of VLMs on this dataset can directly contribute to our understanding of the progress towards the expert vision language model with multilingual capability.

    In future work, we plan to extend the dataset with more image samples, subjects, languages and modalities.

% \todo[inline]{Add something about future work: cen be just 1-2 sentences, e.g., add new languages, more data, ...}

% \section{Conclusions}
    
\section*{Limitations}
Despite its comprehensive nature, \dataset, like any benchmark, is not without limitations. For ease of evaluation and analysis, we only considered and collected multiple-choice questions. We limited our multimodal analysis to four broad categories, which are scientific symbols, figures, graphs, and tabular data. But this can be further extended to finer-grained analysis. For example, scientific symbols can be further broken down into mathematical notions and chemical symbols, while figures can be broken down into maps, figures, paintings, diagrams, etc. However, this requires the collection of more data, which is difficult, particularly for low-resource languages like Croatian, Serbian, and Arabic under consideration. Furthermore, since we are collecting exam questions from different regions of the world, the difficulty of the questions varies depending on the region they originate from. This hurts the comparability of the dataset across languages. Although we tried to include parallel questions for direct comparability, but it was feasible only for three European languages: Croatian, Serbian, and Italian.
\section*{Ethical Consideration}
\begin{itemize}
    \item \textbf{Copyright and Licensing:} All data in \dataset are collected from public sources. 
    \item \textbf{Ethics and Data Privacy:} 
    All testing instances in \dataset are carefully scrutinized to exclude any examples with ethical concerns. Since all the data are collected from exam papers there is no privacy issue.
\end{itemize}

\bibliographystyle{acl_natbib}
\bibliography{custom}

\appendix

\clearpage
\onecolumn

\section{Dataset statistics}
The statistics of the \dataset dataset for all languages and subjects are presented in Table \ref{tab:stat_data2}. Table \ref{tab:test_dist} shows the distribution of multimodal data in the test dataset used for the evaluation of VLMs and LLMs. One point to note is that there are instances where a single image might have multiple modalities, e.g. figure and table. We count them in each of the categories in the table.
\label{sec:data_stats}

\begin{table*}[h!]
\centering
\resizebox{0.9\linewidth}{!}{
\begin{tabular}{lrrrrrrrrrrr}
\toprule
\textbf{Subjects} 	& \textbf{\bul} 	& \textbf{\chin} 	& \textbf{\hrv} 	&\textbf{\srb} 	&\textbf{\itl} 	& \textbf{\fra} 	& \textbf{\deu} 	& \textbf{\hun} 	&  \textbf{\eng} 	& \textbf{\ara}	& \textbf{\pol} \\ 
\midrule
Physics 	& 970 	& 408 	&649 	&305	&215	&235	&510	&1,570	&185 	&67	&\\
Chemistry 	& 665 	& 381 	& 427 	&322	&212	&	&14	&697	& 347	&150 	&\\
Biology 	&233	& 281  	&574 	&294	&424	&	&	&	&47	& 67	&  \\ 
Geography 	& 	&678 	& 383	&54	&40	&24	&46	&92	& 	&	&\\ 
Sociology 	& 264	& 	& 295	&30	&109	&	&	&	&	&306	&\\ 
Business   	& 	& 	& 	&6	&	&180	&216	&747	&  	&	&\\
History 	& 	& 209 	& 500	&200	& 235	&	&	&	& 	&	& \\ 
Philosophy 	& 	& 	&140	&12	&34 	&	&	&	&	&	& \\ 
Psychology  	& 	& 	&154	&47	& 105	&	&	&	&	&	& \\ 
Politics 	& 	& 	&270 	&90	&100	&	&	&	&	&	& \\ 
Informatics 	& 	& 	&188	&74	&146 	&	&	&	& 	&	&\\ 
Mathematics 	& 	&678 	&	&	& 	&	&	&	&145	&	&\\ 
Ethics 	& 	& 	& 180	&	&25	&	&	&	& 	&	& \\ 
Tourism 	& 	& 	& 	&	&	&	&33	&43	&	&	&  \\ 
Science 	& 	& 	& 	&	&	&	&	&	&20	&150	&\\ 
Professional 	& 	& 	& 	&	&	&	&	&	&	& 	&2,511 \\
Islamic Studies 	& 	& 	& 	&	&	&	&	&	&	& 83	&\\ 
Religion 	& 	& 	&161 	&	&	&	&	&	& 	&	&\\ 
Fine Arts 	& 	& 	& 48	&	&	&	&   \\ 
Agriculture 	& 	& 	& 	&	&	&	&	&652	&	&	& \\ 
\midrule
Overall 	& 2,132 	& 2,635	& 3,969	&1,434	&1,645	&439	&819	&3,801	&744	& 823	&2,511\\ 
\bottomrule
\end{tabular}}
\caption{Detailed statistics of \dataset.}
\label{tab:stat_data2}
\end{table*}

\begin{table*}[h!]
\centering
\begin{tabular}{lccccc}
\toprule
Languages & Table & Figure & Graph & Symbol & Text \\ \midrule
English & 0 & 38 & 28 & 43 & 159 \\ %\hline
Chinese & 82 & 246 & 56 & 56 & 126 \\ %\hline
French & 0 & 37 & 6 & 9 & 124 \\ %\hline
German & 0 & 35 & 10 & 8 & 179 \\ %\hline
Italian & 11 & 151 & 13 & 39 & 382 \\ %\hline
Arabic & 11 & 81 & 12 & 29 & 211 \\ %\hline
Polish & 14 & 36 & 0 & 0 & 50 \\ %\hline
Hungarian & 0 & 36 & 7 & 80 & 270 \\ %\hline
Bulgarian & 6 & 70 & 10 & 51 & 200 \\ %\hline
Croatian & 19 & 330 & 24 & 62 & 780 \\ %\hline
Serbian & 11 & 151 & 13 & 39 & 357 \\ \bottomrule
\end{tabular}
\caption{Vision feature distribution of \dataset test set}
\label{tab:test_dist}
\end{table*}

\clearpage
\newpage

\section{Example of OCR and GPT-4V Caption Output}\label{app:ocr_cap}
This section shows examples of OCR and GPT-4V for different vision features.  
\begin{figure*}[h!]
    \centering
    \includegraphics[width=0.9\textwidth]{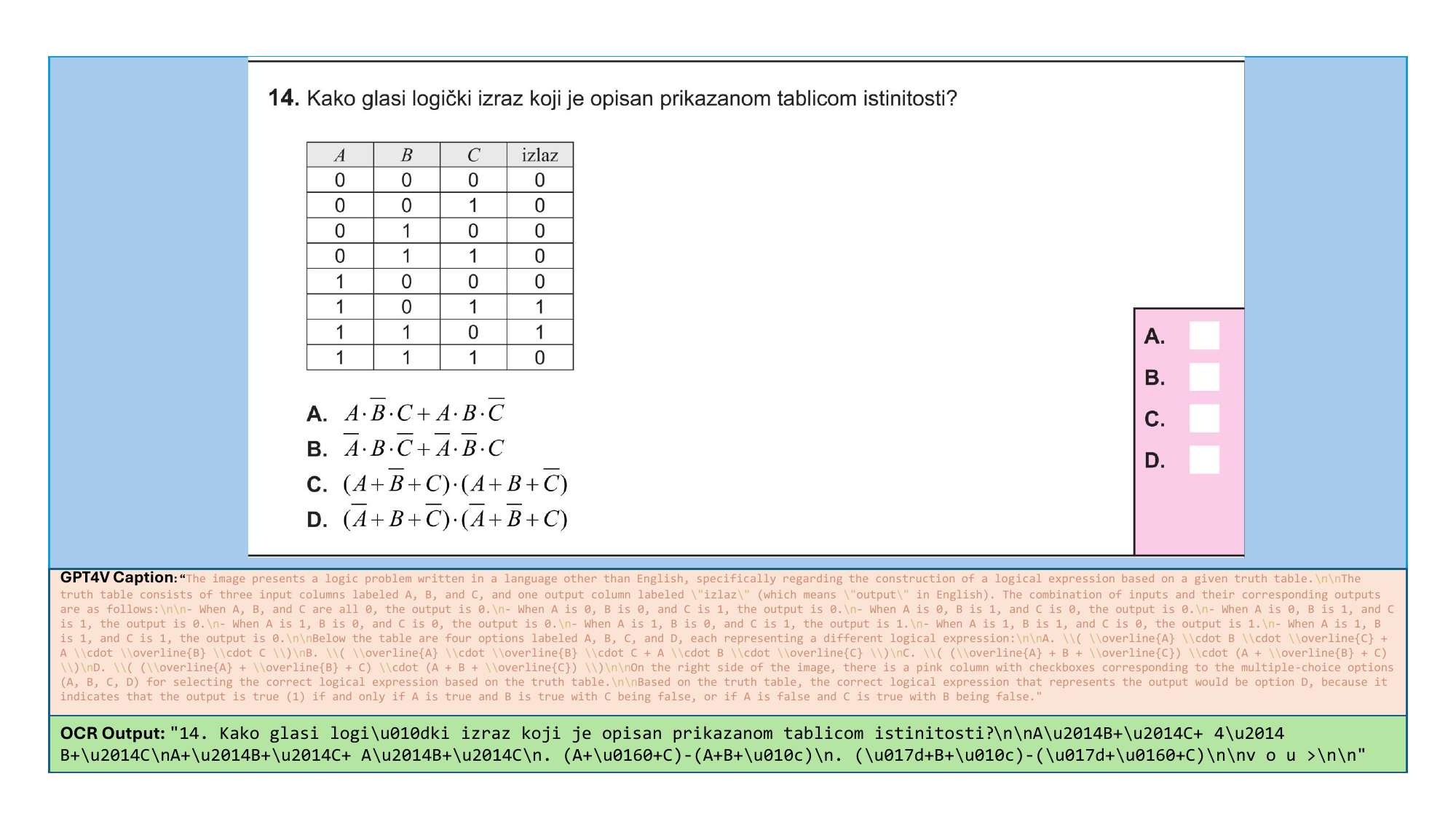}
    \caption{Example of OCR and GPT-4V caption output when provided image with tabular data}
    \label{fig:tab_ocr}
\end{figure*}

\begin{figure*}[h!]
    \centering
    \includegraphics[width=0.9\textwidth]{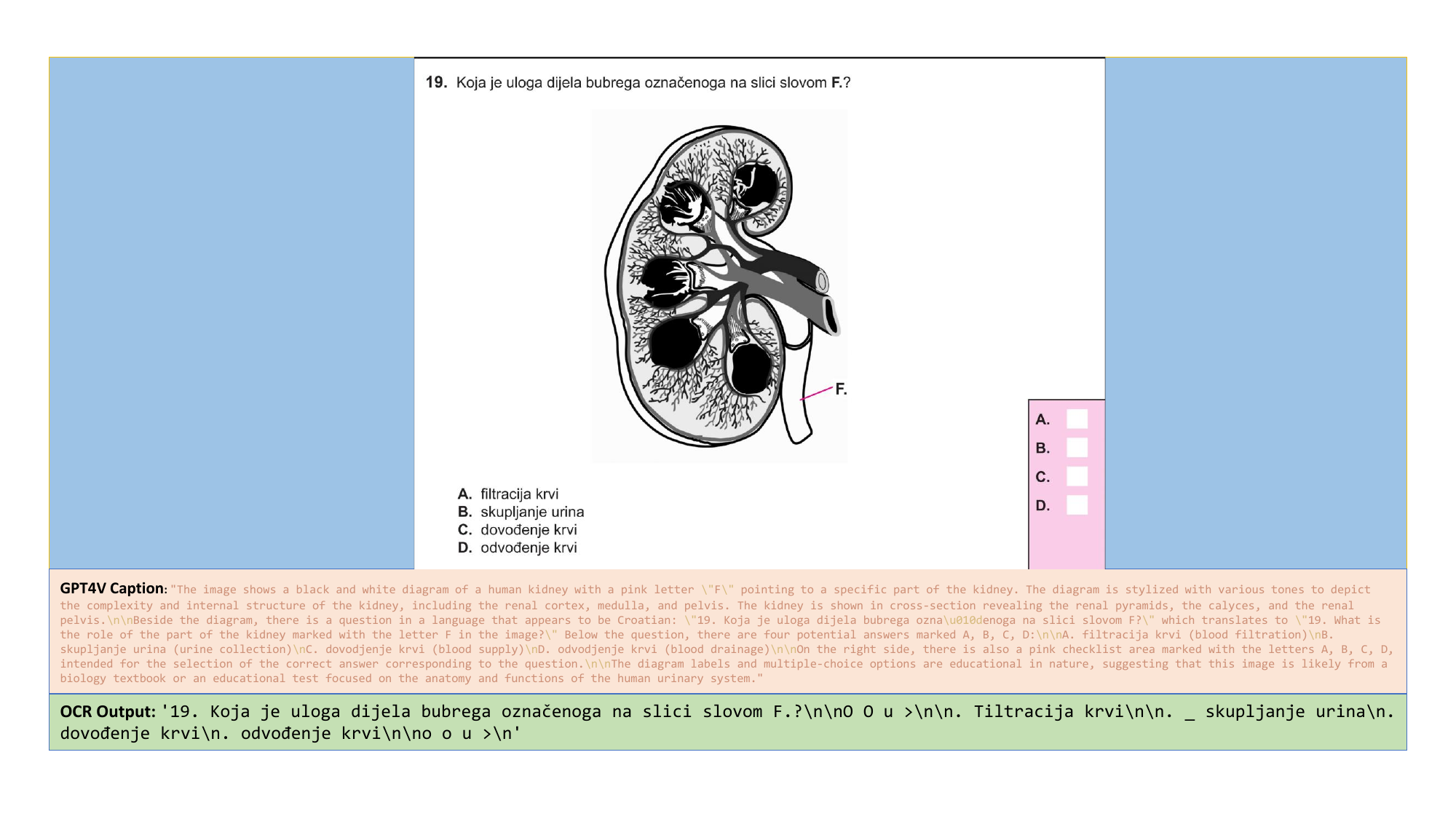}
    \caption{Example of OCR and GPT-4V caption output when provided image with figure}
    \label{fig:image_ocr}
\end{figure*}

\begin{figure*}[h!]
    \centering
    \includegraphics[width=0.9\textwidth]{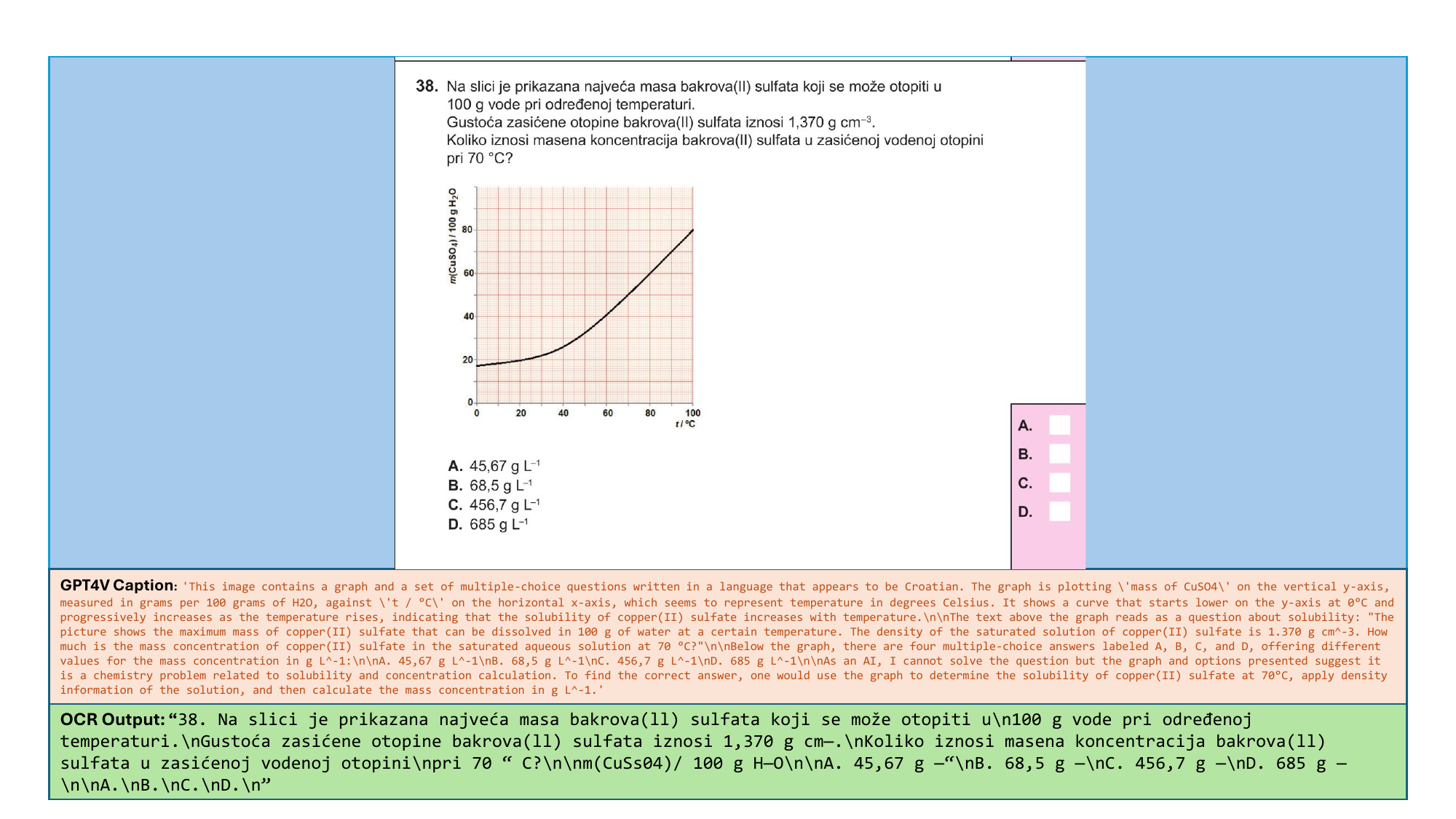}
    \caption{Example of OCR and GPT-4V caption output when provided image with graphical data}
    \label{fig:graph_ocr}
\end{figure*}

\begin{figure*}[h!]
    \centering
    \includegraphics[width=0.9\textwidth]{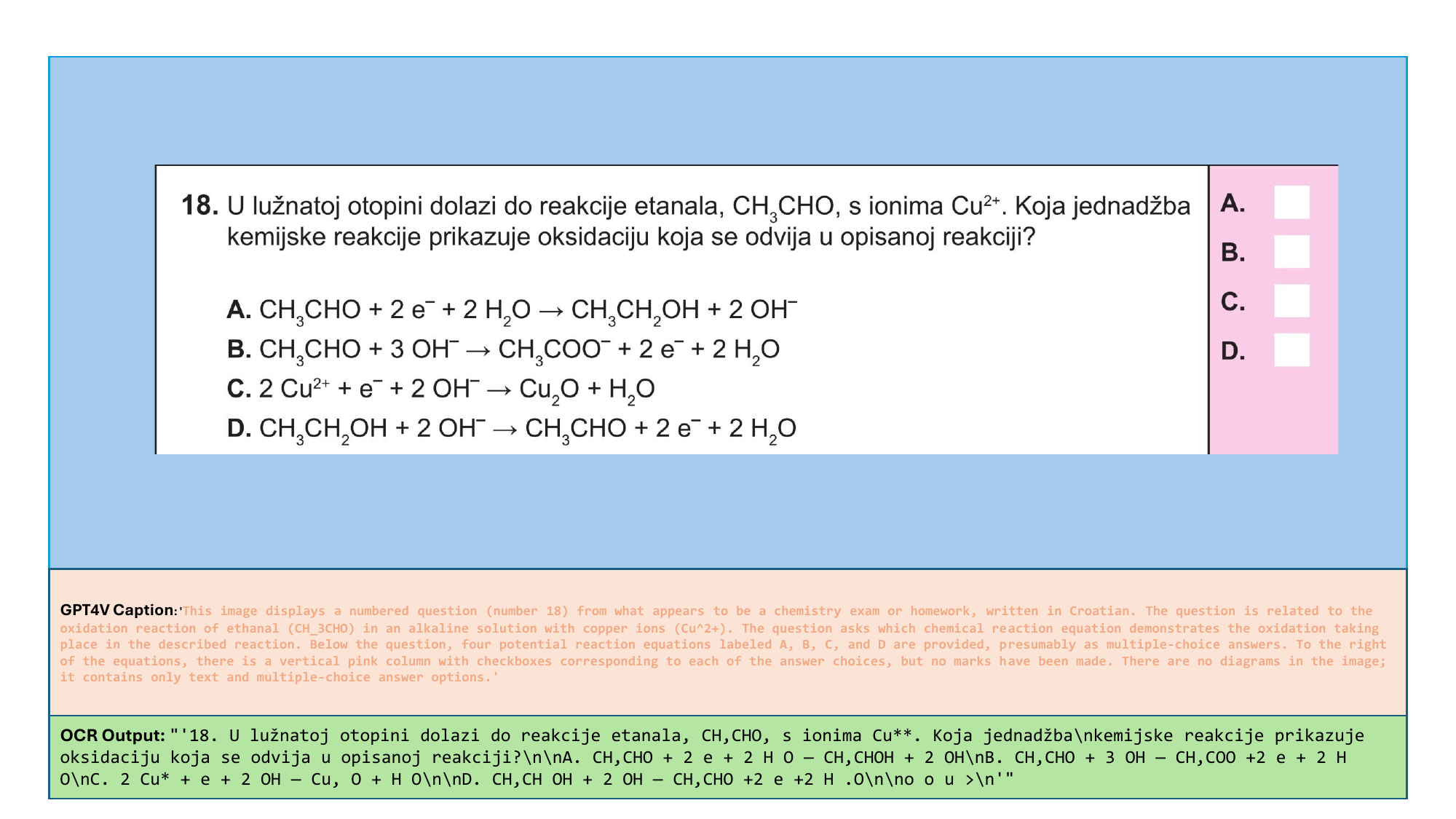}
    \caption{Example of OCR and GPT-4V caption output when provided image with chemistry symbols data}
    \label{fig:chem_ocr}
\end{figure*}

\clearpage
\newpage

\section{Data Quality assessment Guideline}
\label{app:guid}
\begin{figure*}[h!]
    \centering
    \includegraphics[width=0.9\textwidth]{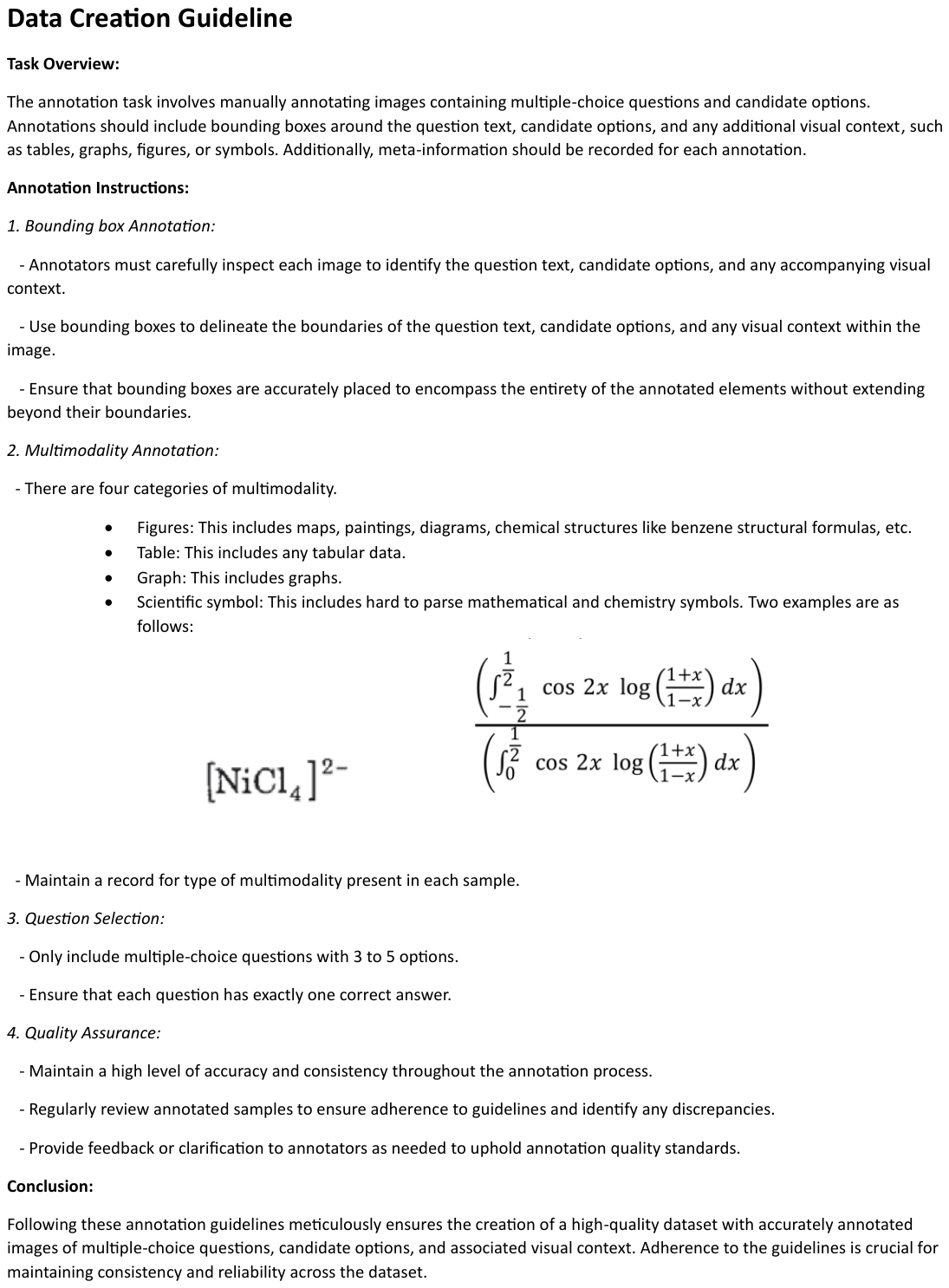}
    \caption{The annotation guideline provided to the annotators while creating the dataset.}
    \label{fig:data_creat_guid}
\end{figure*}

Figure \ref{fig:data_creat_guid} shows a snapshot the annotation guideline shared with the annotators who created the data. The data creation annotators are two authors of the paper, one of which is from India and the other from Bulgaria.

\begin{figure*}[h!]
    \centering
    \includegraphics[width=0.9\textwidth]{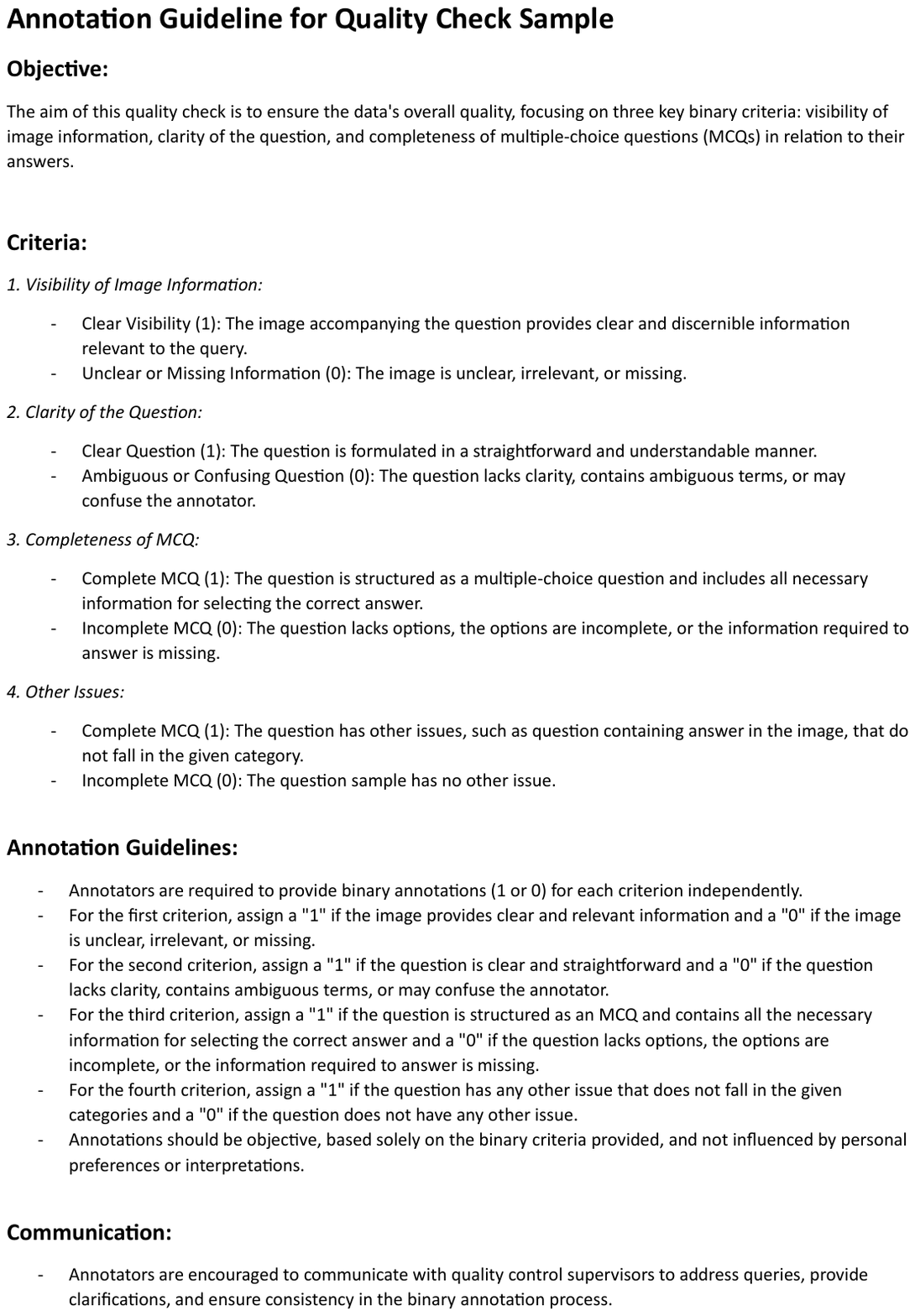}
    \caption{The annotation guideline provided to the annotators to assess the quality of the samples.}
    \label{fig:guideline}
\end{figure*}

Figure \ref{fig:guideline} shows a snapshot of the annotation guideline shared with annotators for quality assessment. The data quality assessment annotators are authors and colleagues with bachelor's degrees and native speakers of the corresponding language.
\clearpage
\newpage

\section{Fine-Grained Evaluation}
\begin{table*}[h!]
  \centering
  \begin{tabular}{lccccc}
    \toprule
    Subject            & GPT-4V & Gemini-Pro-Vision & Gemini-Pro & GPT3.5 & GPT4 \\ \midrule
    Mathematics & 12.00 & 15.0              & 18.00      & 20.00  & 14.00 \\ %\hline
    Chemistry   & 31.00 & 28.0              & 30.00      & 31.00  & 42.00 \\ %\hline
    Physics     & 31.00 & 30.0              & 30.00      & 24.00  & 25.00 \\ %\hline
    \bottomrule
  \end{tabular}
  \caption{Fine-Grained Subject Wise Evaluation of English}
  \label{tab:English-Finegrained}
\end{table*}

\begin{table*}[h!]
  \centering
  \begin{tabular}{lccccc}
    \toprule
    Subject          & GPT-4V & Gemini-Pro-Vision & Gemini-Pro & GPT3.5 & GPT4 \\ \midrule
    Biology   & 22.00 &  0.1700           &  17.00     & 17.00  & 24.00  \\ %\hline
    Chemistry & 30.00 &  0.3200           &  24.00     & 20.00  & 27.00  \\ %\hline
    Geography & 14.00 &  0.2600           &  24.00     & 24.00  & 18.00  \\ %\hline
    History   & 19.00 &  0.2300           &  22.00     & 23.00  & 27.00  \\ %\hline
    Physics   & 22.86 &  0.2429           &  27.14     & 28.57  & 14.29  \\ %\hline
    Science   & 21.35 &  0.2584           &  31.46     & 23.59  & 25.84  \\ %\hline
    \bottomrule
  \end{tabular}
  \caption{Fine-Grained Subject Wise Evaluation of Chinese}
  \label{tab:Chinese-Finegrained}
\end{table*}

\begin{table*}[h!]
  \centering
  \begin{tabular}{lccccc}
    \toprule
     Subject                     & GPT-4V & Gemini-Pro-Vision & Gemini-Pro & GPT3.5 & GPT4 \\ \midrule
    Business \& Economics & 62.00 & 48.00             & 38.00      & 44.00  & 34.00 \\ %\hline
    Geography             & 79.17 & 54.17             & 58.33      & 45.83  & 62.00 \\ %\hline
    Physics               & 55.00 & 26.00             & 34.00      & 35.00  & 32.00 \\ %\hline
    \bottomrule
  \end{tabular}
  \caption{Fine-Grained Subject Wise Evaluation of French}
  \label{tab:French-Finegrained}
\end{table*}

\begin{table*}[h!]
  \centering
  \begin{tabular}{lccccc}
    \toprule
    Subject                      & GPT-4V & Gemini-Pro-Vision & Gemini-Pro & GPT3.5 & GPT4 \\ \midrule
    Business \& Economics & 16.67 & 22.92             & 16.67      & 35.42  & 20.83 \\ %\hline
    Geography             & 78.26 & 56.52             & 41.30      & 30.43  & 21.74 \\ %\hline
    Physics               & 52.00 & 52.00             & 14.00      & 33.00  & 25.00 \\ %\hline
    Tourism               & 63.64 & 60.61             & 30.30      & 45.45  & 27.27 \\ %\hline
    \bottomrule
  \end{tabular}
  \caption{Fine-Grained Subject Wise Evaluation of German}
  \label{tab:German-Finegrained}
\end{table*}

\begin{table*}[h!]
  \centering
  \begin{tabular}{lccccc}
    \toprule
    Subject            & GPT-4V & Gemini-Pro-Vision & Gemini-Pro & GPT3.5 & GPT4 \\ \midrule
    Biology     & 66.90 & 43.41    & 64.73     & 59.39  & 77.55 \\ %\hline
    Chemistry   & 53.33 & 34.67    & 54.67      & 52.00  & 72.00 \\ %\hline
    Ethics      & 84.00 & 36.00    & 96.00      & 80.00  & 100 \\ %\hline
    Geography   & 60.00 & 44.00    & 60.00     & 36.00  & 72.00 \\ %\hline
    History     & 61.73 & 37.04    & 56.79      & 50.62  & 76.54 \\ %\hline
    Informatics & 40.74 & 46.29    & 53.70      & 46.30  & 66.67 \\ %\hline
    Philosophy  & 76.47 & 44.12    & 76.47      & 73.53  & 85.29 \\ %\hline
    Physics     & 51.39 & 31.94    & 51.39      & 34.72  & 62.50 \\ %\hline
    Politics    & 73.33 & 64.44    & 82.22      & 68.69  & 86.67 \\ %\hline
    Psychology  & 88.89 & 59.26    & 85.19      & 77.78  & 92.59 \\ %\hline
    Sociology   & 60.00 & 56.67    & 76.67      & 66.67  & 70.00  \\ %\hline
    \bottomrule
  \end{tabular}
  \caption{Fine-Grained Subject Wise Evaluation of Italian}
  \label{tab:Italian-Finegrained}
\end{table*}

\begin{table*}[h!]
  \centering
  \begin{tabular}{lccccc}
    \toprule
     Subject               & GPT-4V & Gemini-Pro-Vision & Gemini-Pro & GPT3.5 & GPT4 \\ \midrule
    Biology         & 29.82 &  21.05            & 10.52      & 29.82  & 28.07 \\ %\hline
    Chemistry       & 25.67 &  21.62            & 28.38      & 16.22  & 20.27 \\ %\hline
    Islamic Studies & 12.00 &  14.00            & 32.00      & 28.00  & 24.00 \\ %\hline
    Physics         & 16.42 &  16.42            & 25.37      & 22.39  & 31.34 \\ %\hline
    Science         & 34.25 &  26.03            & 27.40      & 26.03  & 31.51 \\ %\hline
    Social          & 24.24 &  15.15            & 15.15      & 37.88  & 23.45 \\ %\hline
    \bottomrule
  \end{tabular}
  \caption{Fine-Grained Subject Wise Evaluation of Arabic}
  \label{tab:Arabic-Finegrained}
\end{table*}

\begin{table*}[h!]
  \centering
  \begin{tabular}{lccccc}
    \toprule
     Subject            & GPT-4V & Gemini-Pro-Vision & Gemini-Pro & GPT3.5 & GPT4  \\ \midrule
    Professional & 30.00 &  28.00            & 42.00     & 33.00  &  30.00 \\ %\hline

    \bottomrule
  \end{tabular}
  \caption{Fine-Grained Subject Wise Evaluation of Polish}
  \label{tab:Polish-Finegrained}
\end{table*}

\begin{table*}[h!]
  \centering
  \begin{tabular}{lccccc}
    \toprule
     Subject                     & GPT-4V & Gemini-Pro-Vision & Gemini-Pro & GPT3.5 & GPT4 \\ \midrule
    Business \& Economics & 37.14 & 37.14             & 52.86      & 41.43  & 45.71     \\ %\hline
    Geography             & 44.00 & 26.00             & 54.00      & 46.00  & 42.00     \\ %\hline
    Physics               & 52.00 & 28.00             & 42.00      & 45.00  & 34.00     \\ %\hline
    Tourism               & 60.47 & 25.58             & 55.81      & 41.86  & 32.56     \\ %\hline
    Landscaping           & 40.74 & 22.22             & 44.44      & 33.33  & 29.63     \\ %\hline
    Chemistry             & 34.00 & 23.00             & 30.00      & 25.00  & 28.00     \\ %\hline
    Agriculture           & 52.00 & 24.00             & 38.00      & 34.00  & 26.00     \\ %\hline
    \bottomrule
  \end{tabular}
  \caption{Fine-Grained Subject Wise Evaluation of Hungarian}
  \label{tab:Hungarian-Finegrained}
\end{table*}

\begin{table*}[h!]
  \centering
  \begin{tabular}{lccccc}
    \toprule
    Subject   & GPT-4V & Gemini-Pro-Vision & Gemini-Pro & GPT3.5 & GPT4 \\ \midrule
    Biology   & 42.67 & 24.00             & 42.67      & 30.67  & 29.33 \\ %\hline
    Chemistry & 35.00 & 27.00             & 43.00      & 23.00  & 27.00 \\ %\hline
    Physics   & 28.00 & 32.00             & 14.00      & 25.00  & 33.00 \\ %\hline
    Sociology & 44.00 & 44.00             & 30.00      & 34.00  & 34.00 \\ %\hline
    \bottomrule
  \end{tabular}
  \caption{Fine-Grained Subject Wise Evaluation of Bulgarian}
  \label{tab:Bulgarian-Finegrained}
\end{table*}

\begin{table*}[h!]
  \centering
  \begin{tabular}{lccccc}
    \toprule
     Subject           & GPT-4V & Gemini-Pro-Vision & Gemini-Pro & GPT3.5 & GPT4 \\ \midrule
    Biology     & 72.04 & 32.26  & 0.4500 & 62.36 & 75.27 \\ %\hline
    Chemistry   & 48.00 &  25.33 & 0.5000 & 48.00 & 72.00 \\ %\hline
    Ethics      & 76.00 & 24.00  & 0.5000 & 84.00 & 100 \\ %\hline
    Fine Arts   & 41.30 &  41.30 & 0.3261 & 36.96 & 47.83 \\ %\hline
    Geography   & 46.93 &  20.04 & 0.3100 & 34.02 & 30.99 \\ %\hline
    History     & 59.26 &  29.63 & 0.5000 & 51.85 & 85.19 \\ %\hline
    Informatics & 38.89 &  42.59 & 0.4940 & 50.00 & 62.96 \\ %\hline
    Philosophy  & 70.59 &  00.00 & 0.6600 & 67.65 & 88.24 \\ %\hline
    Physics     & 45.84 &  27.78 & 0.3700 & 40.28 & 61.11 \\ %\hline
    Politics    & 82.22 &  46.67 & 0.3200 & 82.22 & 97.78 \\ %\hline
    Psychology  & 85.19 &  33.33 & 0.5200 & 81.48 & 92.59 \\ %\hline
    Religion    & 26.00 &  28.00 & 0.5600 & 28.00 & 30.00 \\ %\hline
    Sociology   & 63.33 &  33.33 & 0.6000 & 70.00 & 80.00 \\ %\hline
    \bottomrule
  \end{tabular}
  \caption{Fine-Grained Subject Wise Evaluation of Croatian}
  \label{tab:Croatian-Finegrained}
\end{table*}

\begin{table*}[h!]
  \centering
  \begin{tabular}{lccccc}
    \toprule
     Subject           & GPT-4V & Gemini-Pro-Vision & Gemini-Pro & GPT3.5 & GPT4 \\ \midrule
    Biology     & 37.64 & 31.18             &  69.89     &  51.61 & 73.12\\ %\hline
    Chemistry   & 28.00 & 26.67             &  52.00     &  38.67 & 68.00\\ %\hline
    Geography   & 36.00 & 32.00             &  72.00     &  40.00 & 80.00\\ %\hline
    History     & 45.68 & 23.46             &  55.56     &  46.91 & 77.78\\ %\hline
    Informatics & 33.33 & 33.34             &  50.00     &  48.15 & 57.41\\ %\hline
    Physics     & 38.89 & 23.62             &  38.39     &  38.89 & 63.89\\ %\hline
    Politics    & 46.67 & \textbf{31.11}    &  77.78     &  60.00 & 91.11\\ %\hline
    Psychology  & 55.56 & \textbf{29.63}    &  85.19     &  62.96 & 100 \\ %\hline
    Sociology   & 53.33 & \textbf{30.00}    &  70.00     &  70.00 & 73.33 \\ %\hline
    \bottomrule
  \end{tabular}
  \caption{Fine-Grained Subject Wise Evaluation of Serbian}
  \label{tab:Serbian-Finegrained}
\end{table*}

\clearpage
\newpage

\section{Sample Example from Different languages}
\begin{figure*}[h!]
    \centering
    \includegraphics[width=0.8\textwidth]{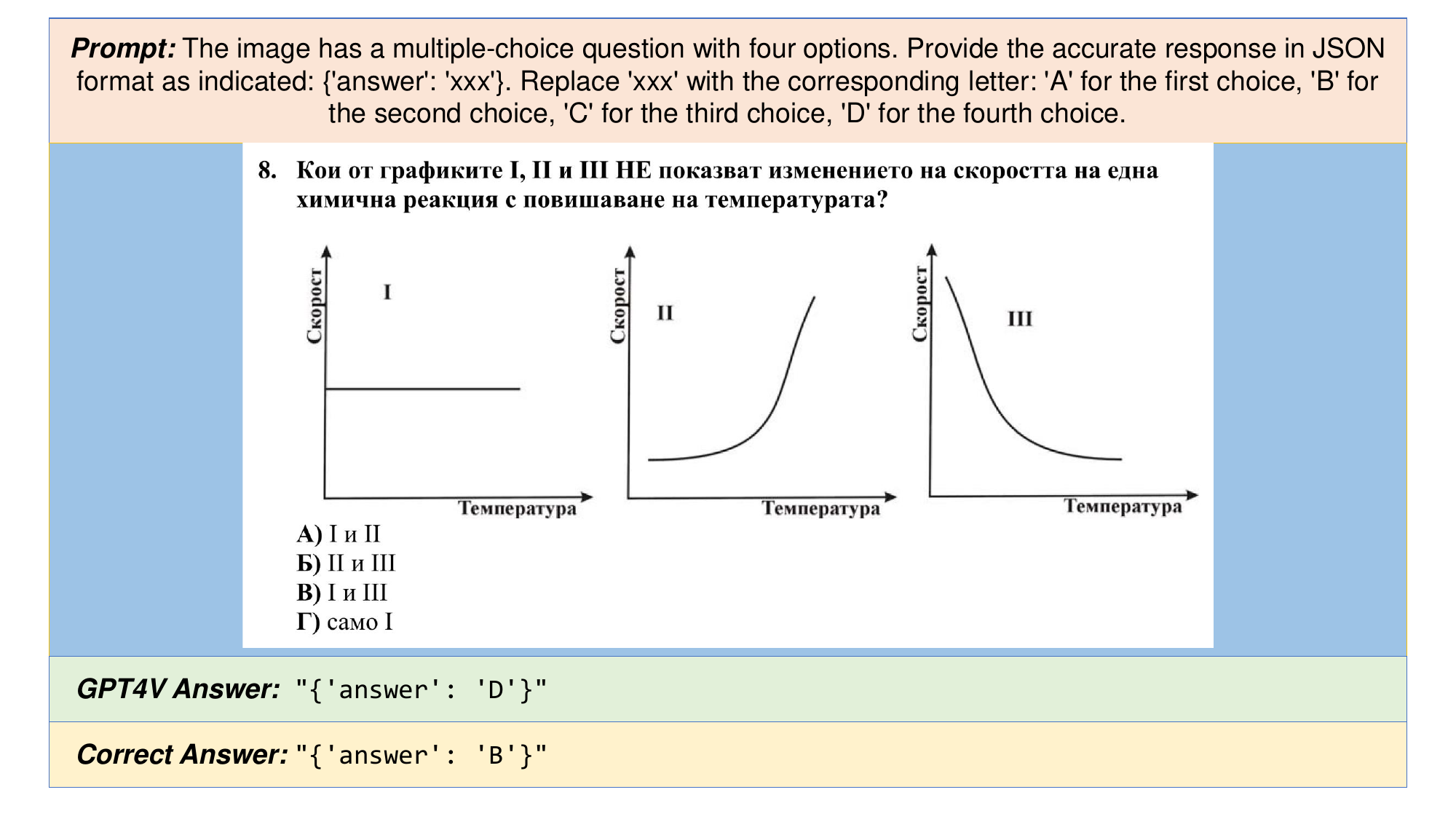}
    \caption{A sample from Bulgarian test set with GPT-4V output}
    \label{fig:bg}
\end{figure*}

\begin{figure*}[h!]
    \centering
    \includegraphics[width=0.8\textwidth]{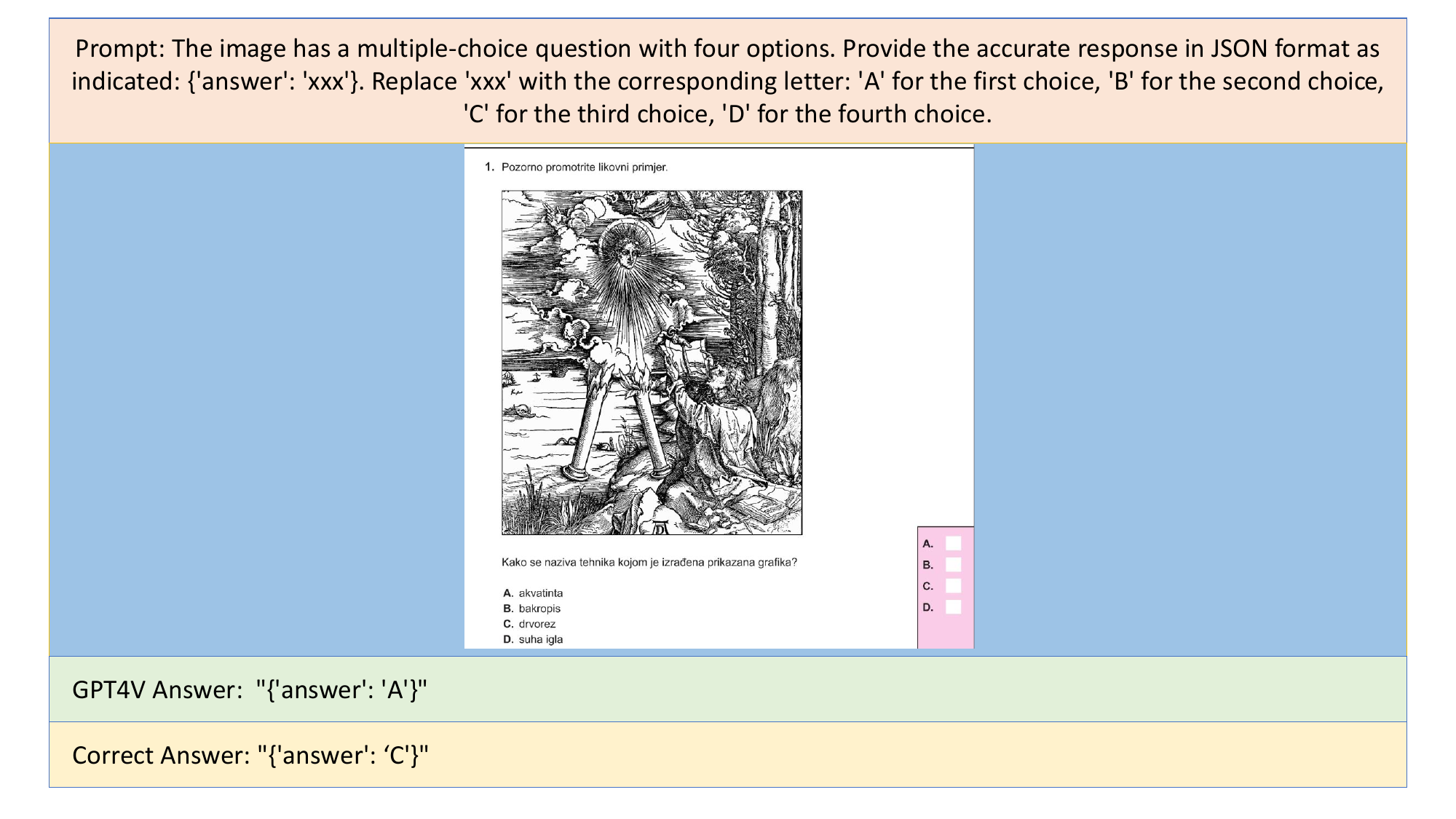}
    \caption{A sample from Croatian test set with GPT-4V output}
    \label{fig:hr}
\end{figure*}

\begin{figure*}[h!]
    \centering
    \includegraphics[width=0.8\textwidth]{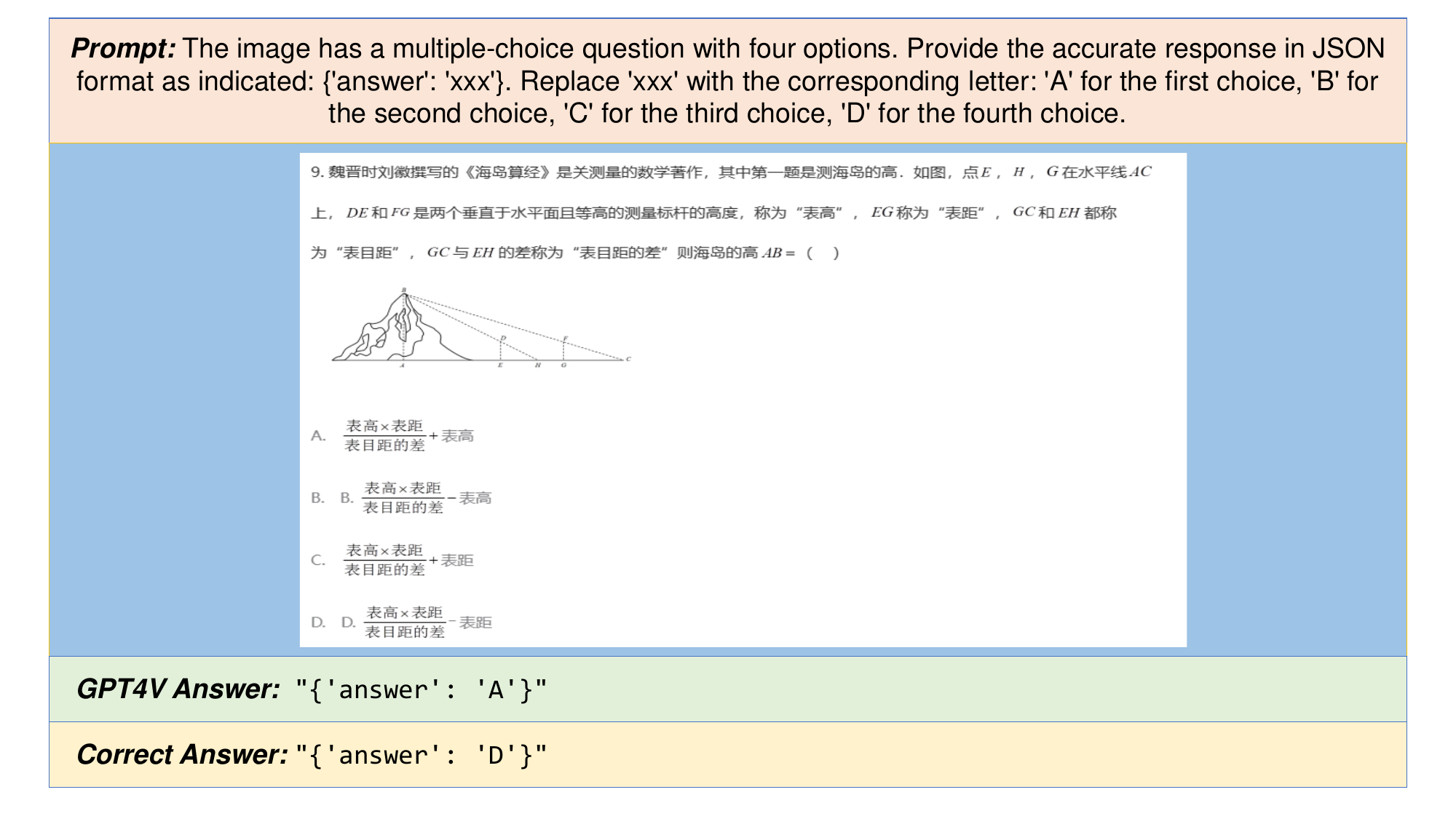}
    \caption{A sample from Chinese test set with GPT-4V output}
    \label{fig:zh}
\end{figure*}

\begin{figure*}[h!]
    \centering
    \includegraphics[width=0.8\textwidth]{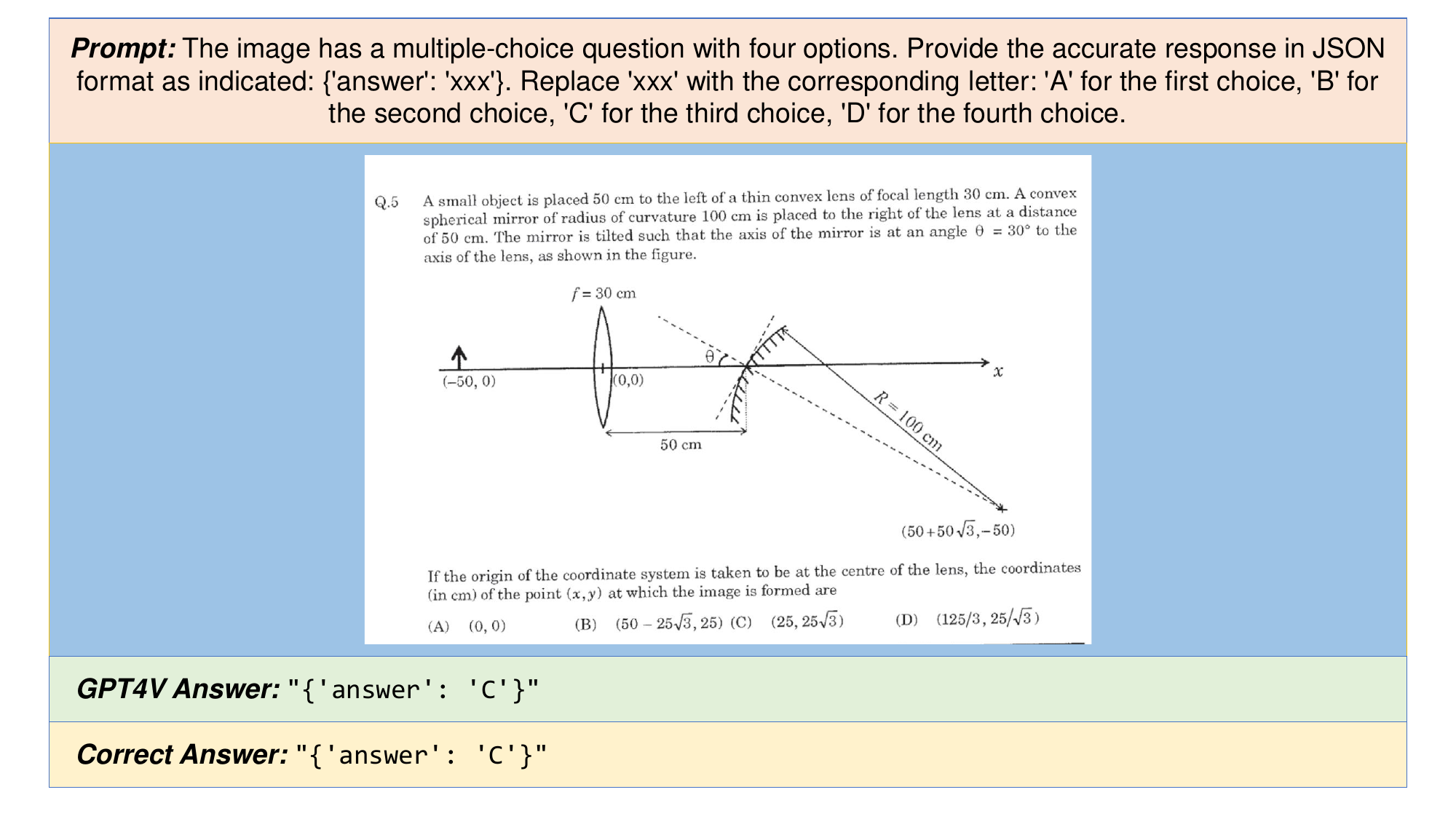}
    \caption{A sample from English test set with GPT-4V output}
    \label{fig:en}
\end{figure*}

\begin{figure*}[h!]
    \centering
    \includegraphics[width=0.8\textwidth]{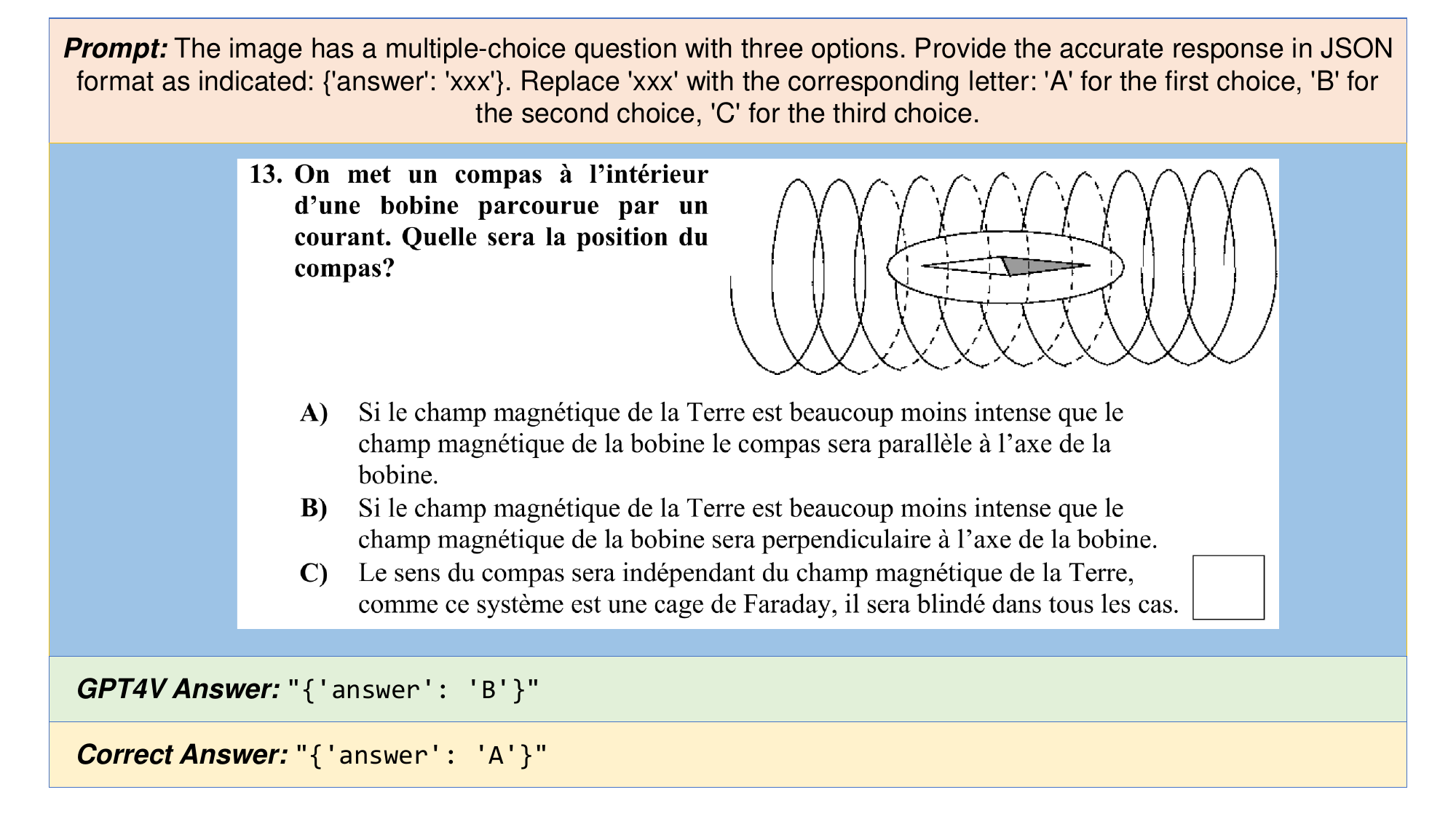}
    \caption{A sample from French test set with GPT-4V output}
    \label{fig:fr}
\end{figure*}

\begin{figure*}[h!]
    \centering
    \includegraphics[width=0.8\textwidth]{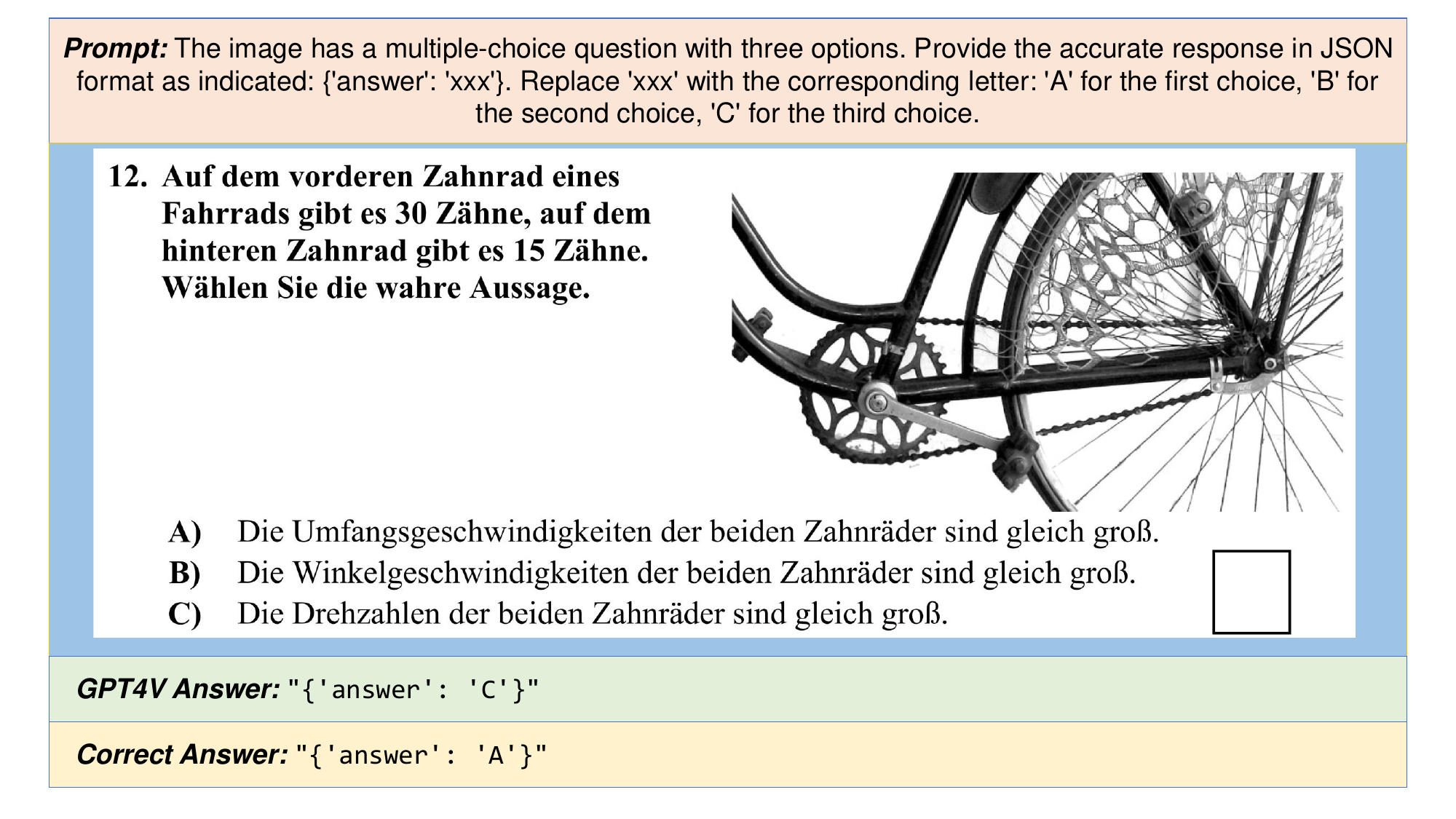}
    \caption{A sample from German test set with GPT-4V output}
    \label{fig:de}
\end{figure*}

\begin{figure*}[h!]
    \centering
    \includegraphics[width=0.8\textwidth]{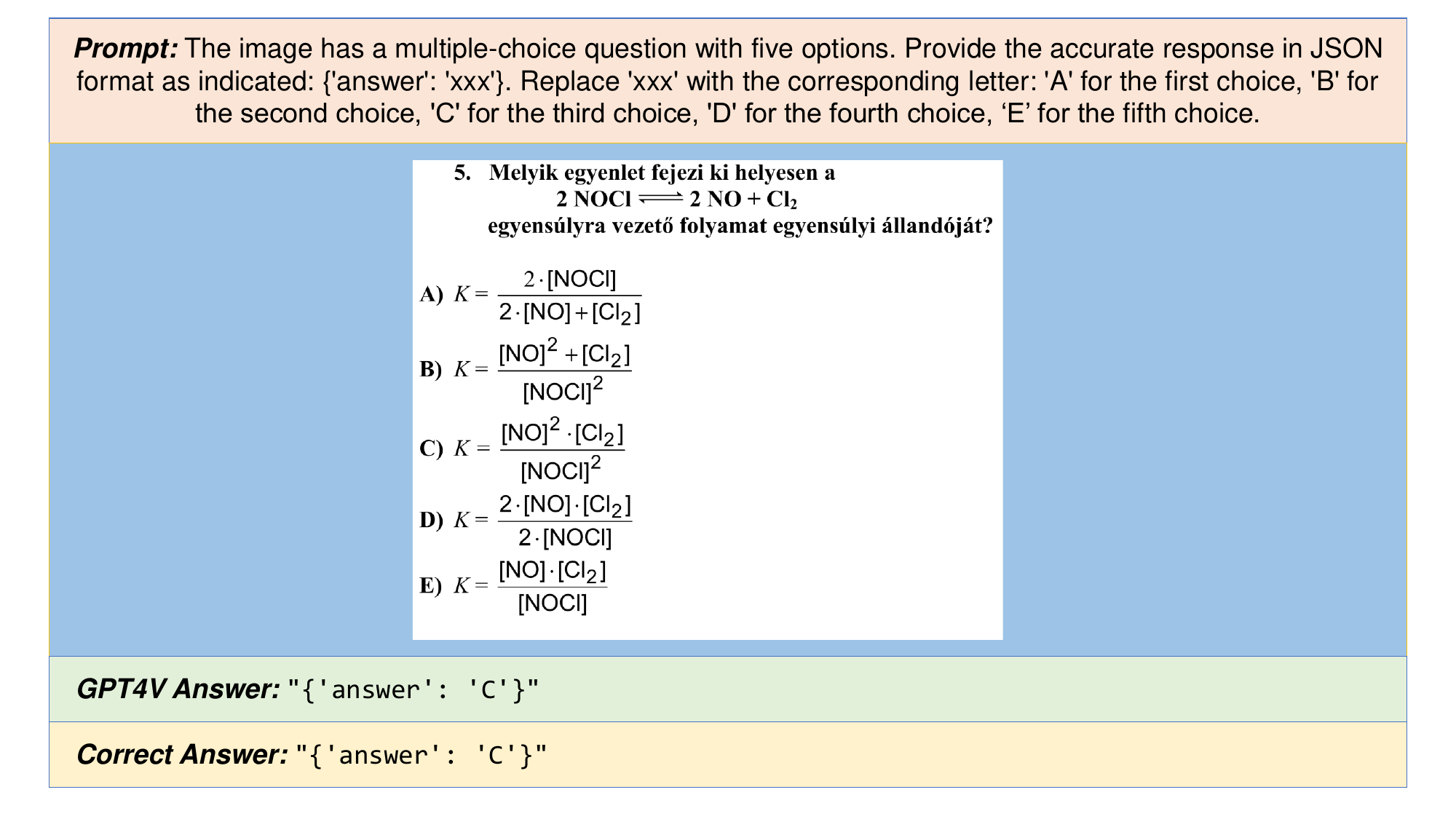}
    \caption{A sample from Hungarian test set with GPT-4V output}
    \label{fig:hu}
\end{figure*}

\begin{figure*}[h!]
    \centering
    \includegraphics[width=0.8\textwidth]{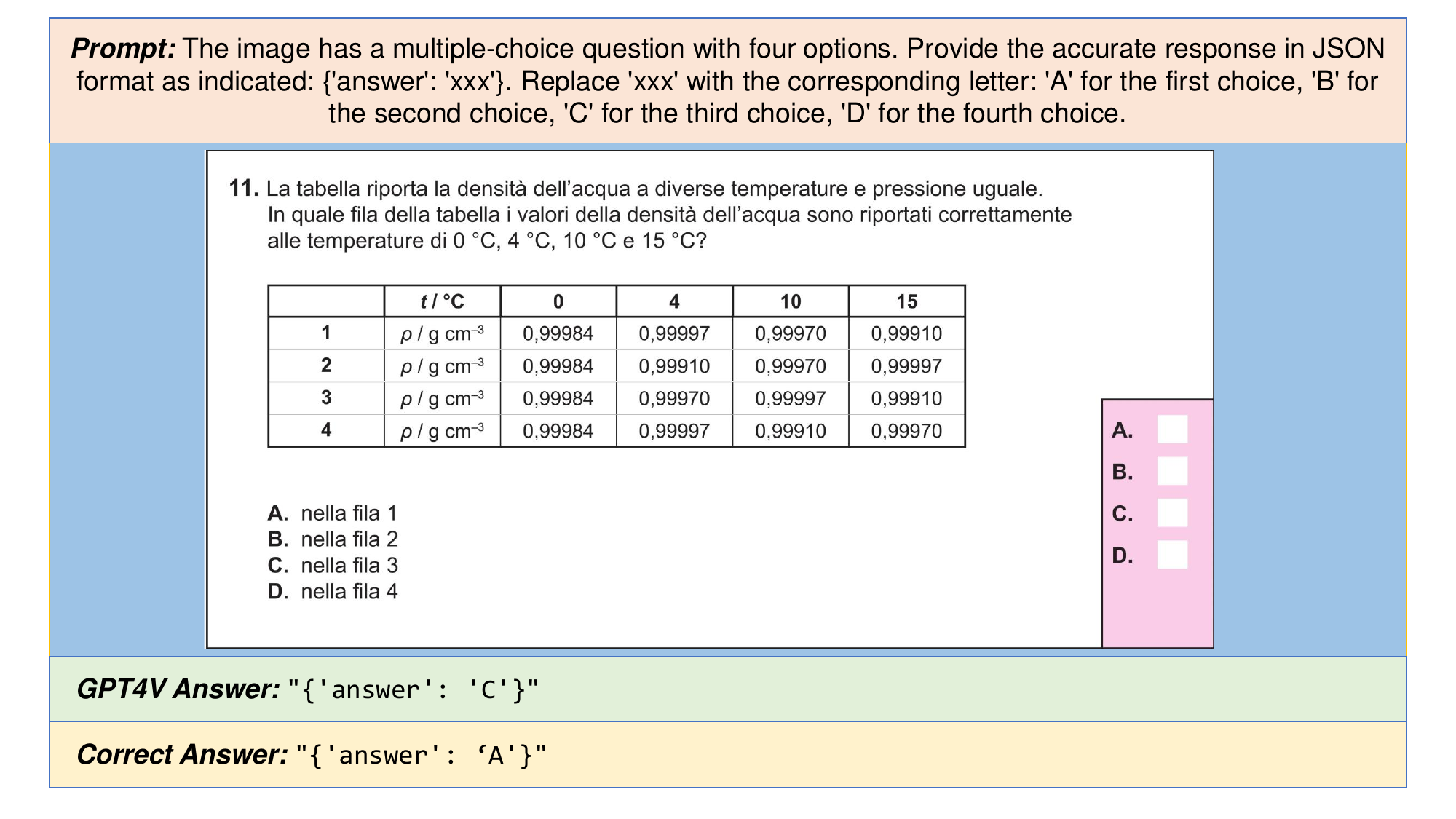}
    \caption{A sample from Italian test set with GPT-4V output}
    \label{fig:it}
\end{figure*}

\begin{figure*}[h!]
    \centering
    \includegraphics[width=0.8\textwidth]{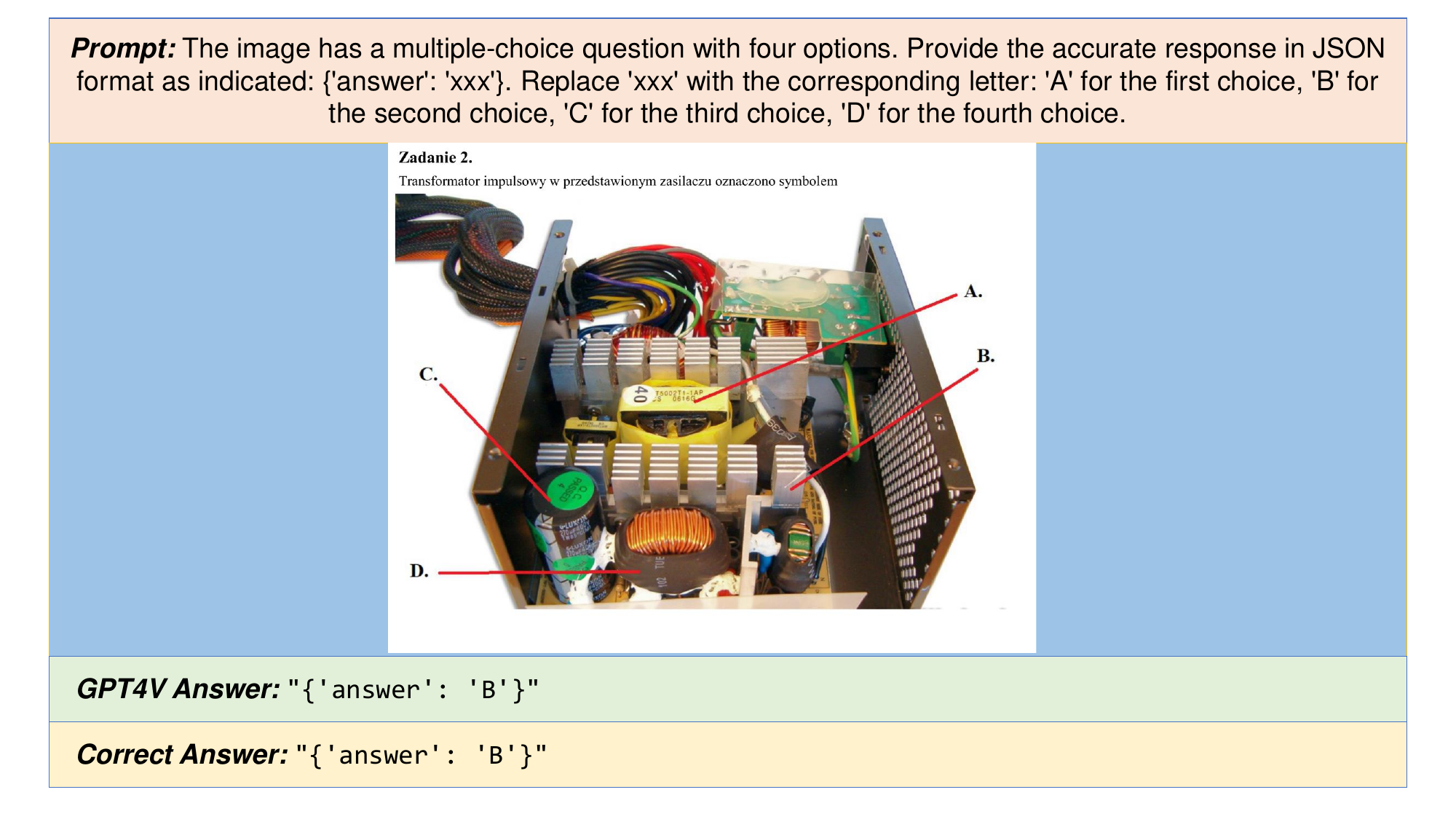}
    \caption{A sample from Polish test set with GPT-4V output}
    \label{fig:pl}
\end{figure*}

\begin{figure*}[h!]
    \centering
    \includegraphics[width=0.8\textwidth]{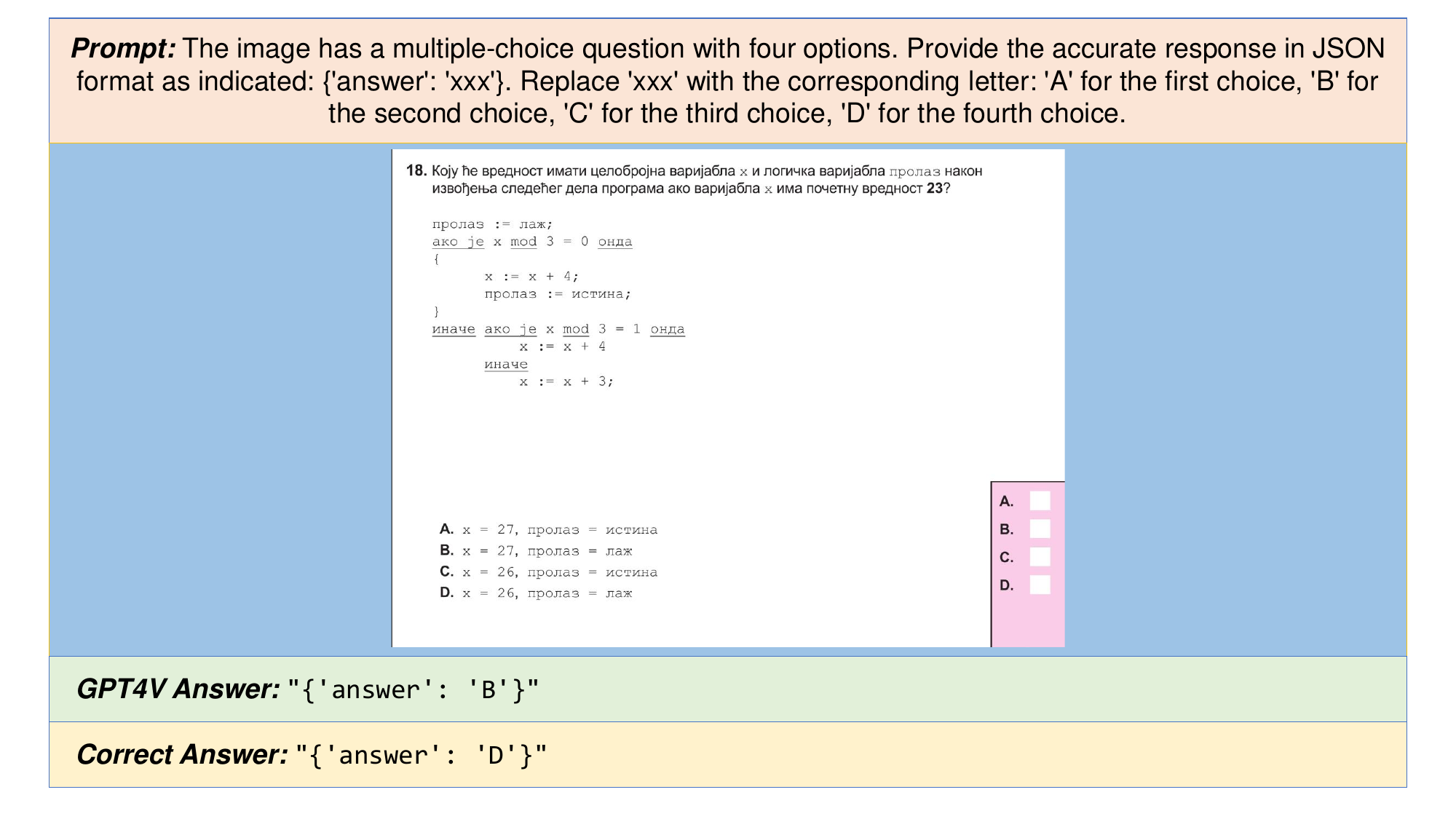}
    \caption{A sample from Serbian test set with GPT-4V output}
    \label{fig:sr}
\end{figure*}

Figure \ref{fig:bg} to \ref{fig:sr} shows sample examples from different languages in \dataset.
\begin{figure*}[h!]
    \centering
    \includegraphics[width=0.8\textwidth]{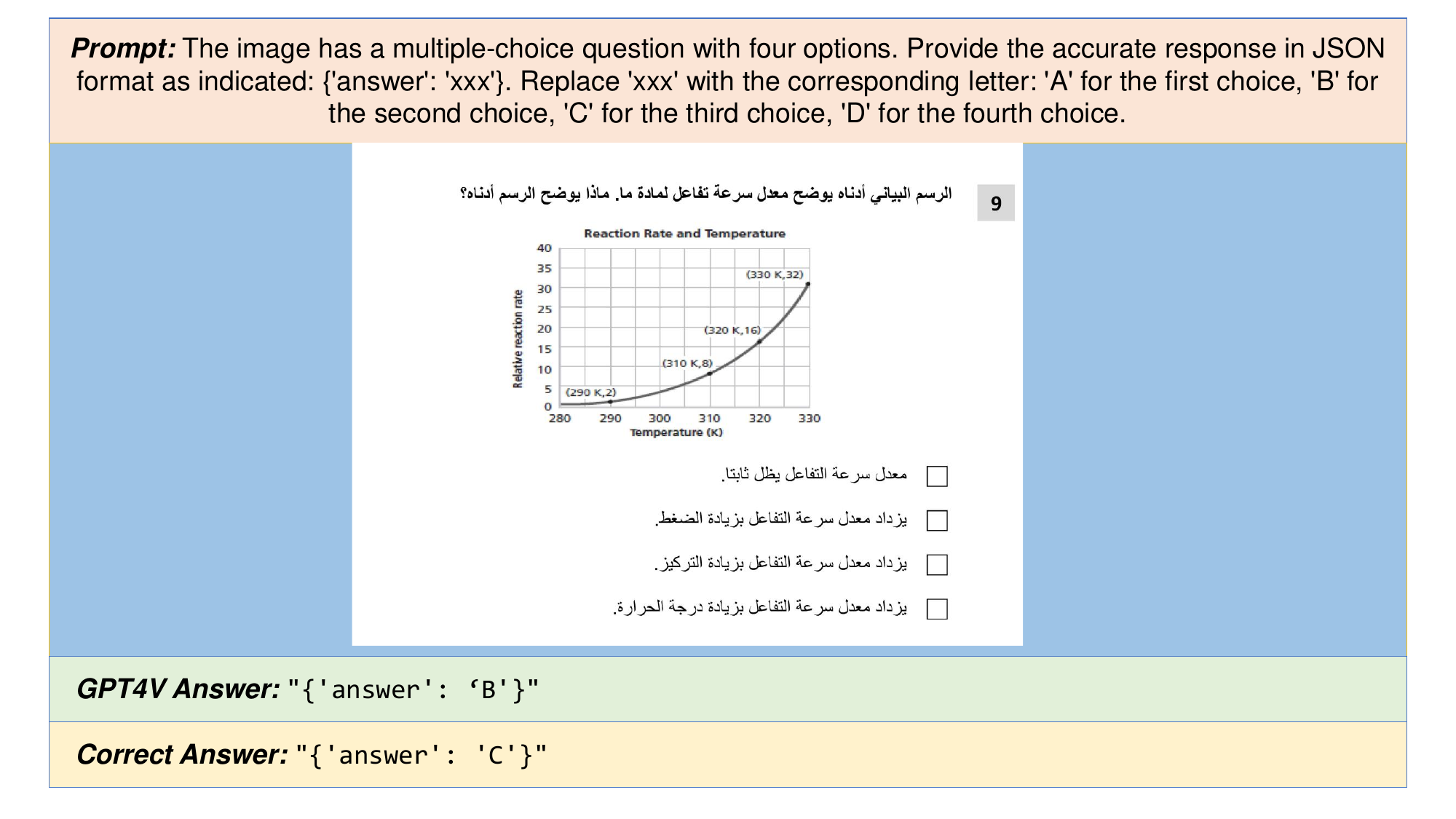}
    \caption{Example from the Arabic test set with a GPT-4V output.}
    \label{fig:ar}
\end{figure*}

% This is an appendix.

\end{document}